\documentclass[twoside,twocolumn,twocolumn]{IEEEtran}
\usepackage[T1]{fontenc}
\usepackage{xcolor}
\usepackage{float}
\usepackage{amsmath}
\usepackage{amsthm}
\usepackage{amssymb}
\usepackage{graphicx}

\usepackage{booktabs} % For better looking tables
\usepackage{makecell} % For multi-line cell titles
\usepackage{tabularx} % For tables that fit the text width

\makeatletter

\floatstyle{ruled}
\newfloat{algorithm}{tbp}{loa}
\providecommand{\algorithmname}{Algorithm}
\floatname{algorithm}{\protect\algorithmname}

  \theoremstyle{plain}
  
  \theoremstyle{plain}
  
  \theoremstyle{plain}

\ifCLASSOPTIONcompsoc
\usepackage[caption=false,font=normalsize,labelfont=sf,textfont=sf]{subfig}
\else
\usepackage[caption=false,font=footnotesize]{subfig}
\fi

\usepackage{cite}
\usepackage{bm}
\usepackage{algorithmic}
\usepackage{algorithm}
\usepackage{graphicx}
\usepackage{bbm}
\usepackage{mathrsfs}
\usepackage{comment}
\IEEEoverridecommandlockouts
\makeatother

\providecommand{\propositionname}{Proposition}
\providecommand{\corollaryname}{Corollary}
\providecommand{\theoremname}{Theorem}

\begin{document}

\title{Video Object Recognition in Mobile Edge Networks: Local Tracking or Edge Detection?}
\author{Kun~Guo,~\IEEEmembership{Member,~IEEE}, Yun~Shen, Xijun~Wang,~\IEEEmembership{Member,~IEEE}, Chaoqun~You,~\IEEEmembership{Member,~IEEE}, \\Yun~Rui,~\IEEEmembership{Senior Member,~IEEE}, and~Tony~Q. S.~Quek,~\IEEEmembership{Fellow,~IEEE} 
\thanks{A preliminary version of this paper has been accepted for presentation at IEEE/CIC ICCC 2025 \cite{ICCC_Yun}.}
\thanks{K. Guo, Y. Shen, and Y. Rui are with the School of Communications and Electronics Engineering, East China Normal University, Shanghai 200241, China (e-mail: kguo@cee.ecnu.edu.cn, 52285904009@stu.ecnu.edu.cn, yrui@ce.ecnu.edu.cn)}
\thanks{X. Wang is with the School of Electronics and Information Technology, Sun Yat-sen University, Guangzhou 510006, China (e-mail: wangxijun@mail.sysu.edu.cn) \textit{(Corresponding author: Xijun Wang.)}}
\thanks{C. You is with the School of Computer Science, Fudan University, Shanghai 200438, China (e-mail: chaoqunyou@gmail.com)}
\thanks{T. Q. S. Quek is with the Information Systems Technology
and Design Pillar, Singapore University of Technology and Design, Singapore 487372 (e-mail: tonyquek@sutd.edu.sg).}
}

\maketitle
\begin{abstract}
Fast and accurate video object recognition, which relies on frame-by-frame video analytics, remains a challenge for resource-constrained devices such as traffic cameras. Recent advances in mobile edge computing have made it possible to offload computation-intensive object detection to edge servers equipped with high-accuracy neural networks, while lightweight and fast object tracking algorithms run locally on devices. This hybrid approach offers a promising solution but introduces a new challenge: deciding when to perform edge detection versus local tracking. To address this, we formulate two long-term optimization problems for both single-device and multi-device scenarios, taking into account the temporal correlation of consecutive frames and the dynamic conditions of mobile edge networks. Based on the formulation, we propose the LTED-Ada in single-device setting, a deep reinforcement learning-based algorithm that adaptively selects between local tracking and edge detection, according to the frame rate as well as recognition accuracy and delay requirement. In multi-device setting, we further enhance LTED-Ada using federated learning to enable collaborative policy training across devices, thereby improving its generalization to unseen frame rates and performance requirements. Finally, we conduct extensive hardware-in-the-loop experiments using multiple Raspberry Pi 4B devices and a personal computer as the edge server, demonstrating the superiority of LTED-Ada.

\end{abstract}

\section{Introduction}
The proliferation of camera deployments in domains such as urban safety and traffic flow monitoring has generated an unprecedented volume of video data \cite{Multi_Cameras, LargeScale, ContinualLearning}. To extract meaningful insights from these data, video object recognition, built upon object recognition across consecutive frames, has emerged as a critical technology \cite{S&T, MOT_MTT, TWC_Offloading}. There are primarily two types of object recognition technologies: object detection and object tracking. In object detection, a pre-trained neural network model is used to process video frames, offering high recognition accuracy at the cost of increased latency and computational load. In contrast, object tracking leverages temporal redundancy between consecutive frames and is considered a promising solution for fast, low-latency object recognition. However, tracking-based methods may suffer from degraded accuracy over time or in the presence of rapid scene changes. 

To harness the complementary strengths of object detection and tracking, edge-assisted video object recognition has emerged as a promising solution. In this paradigm, computation-intensive object detection is offloaded from the device to edge servers via mobile edge networks \cite{TWC_Zhangjun,TAO_MM,Dynamic_MEC}, while lightweight object tracking is performed locally on the device. To be more specific, edge detection is applied to keyframes that exhibit significant changes or rich visual content, whereas local tracking propagates object information from the most recent keyframe across intermediate frames. This brings forth a critical question: how to select the keyframes for fast and accurate video object recognition?
% how can keyframes be effectively selected to balance recognition accuracy and delay? 
In other words, when should edge detection be triggered, and when is local tracking sufficient, with respect to the temporal correlation of consecutive frames and the dynamic conditions of mobile edge networks?

To this end, we formulate two long-term optimization problems that account for the dynamic frame arrivals, network conditions, and queuing management. One is for the single-device scenario, and the other for the multi-device scenario, with the achievable reward for each frame defined as a weighted sum of recognition accuracy, handling delay, and waiting delay. To solve the single-device problem, we propose a deep reinforcement learning (DRL)-based video object recognition algorithm named LTED-Ada, which intelligently selects between \underline{l}ocal \underline{t}racking and \underline{e}dge \underline{d}etection, \underline{ada}pting to the frame rates as well as recognition accuracy and delay requirements. For the multi-device case, we incorporate federated learning into LTED-Ada, enabling its generalization capability on unseen frame rates and performance requirements.  
% enabling it to adapt to varying device requirements on recognition accuracy and delay. 
Finally, we conduct extensive hardware-in-the-loop experiments to demonstrate the superiority of LTED-Ada.

The main contributions of this paper are threefold and can be summarized as follows:
\begin{itemize}
    \item \textbf{Frame Rate Adaptation via Queue Awareness:} We model the queuing systems on both the device and the edge server to account for varying frame rates. This enables precise delay calculation for each frame, regardless of whether it is processed with local tracking or offloaded for edge detection. By incorporating queue lengths as a part of the states in the DRL-based decision-making process, the proposed LTED-Ada dynamically adapts to different frame rates.
    \item \textbf{Generalization Enhancement through Multi-Device Collaboration:} To improve the generalization capability of LTED-Ada, we incorporate federated learning to enable collaborative policy learning across multiple devices. This allows the employed Deep Q-Network (DQN) model on each device to learn from a broader set of state-action pairs, enhancing its robustness to unseen frame rates and performance requirements.
    \item \textbf{Hardware-in-the-Loop Experiments for Realistic Environment Validation:} We conduct extensive experiments using multiple Raspberry Pi 4B devices and a personal computer (PC) serving as the edge server to emulate realistic mobile edge computing environments. Experimental results demonstrate that the LTED-Ada outperforms multiple baselines by effectively balancing recognition accuracy and delay across a wide range of frame rates and performance requirements.
\end{itemize}

The rest of this paper is organized as follows: In Section \ref{sec: related_work}, we summarize the related works. In Section \ref{sec: problem_description}, we give the system model. In Sections \ref{sec: single-device-alg} and \ref{sec: multi-device-alg}, we formulate the long-term optimization problem and propose the LTED-Ada in single- and multi-device scenario, respectively. Experimental results are showcased in Section \ref{sec: results}. Finally, we draw conclusions in Section \ref{sec: conclusions}.

\section{Related Works\label{sec: related_work}}
In this section, we review mainstream research in the field of video object recognition, which can be categorized into two groups based on frame rates. The first targets light-load scenarios, characterized by lower frame rates and no frame stacking. The second focuses on heavy-load scenarios, involving high frame rates and frame queuing.

\subsection{Methods for Light-load Scenarios}
The proposed approaches for dynamically selecting between edge detection and local tracking primarily fall into two categories: threshold-based methods and direct decision methods. In threshold-based methods, a frame is offloaded to the edge server for object detection when a predefined threshold is triggered; otherwise, it is processed locally via object tracking \cite{Glimpse,9969129, 9488741, TMC_SongtaoGuo}. In contrast, direct decision methods cast the offloading decision as a binary variable within an optimization problem, which is then solved to determine the decision \cite{ICASSP_MultiSever,TMC_Min}.

In the threshold-based category, Glimpse, a continuous, real-time object recognition system, proposed in \cite{Glimpse}, switched between neural network-based detection at the edge and optical flow-based tracking locally, based on a fixed pixel deviation threshold between the current frame and the most recent keyframe. More adaptive approaches have also been explored. For instance, \cite{9969129} employed a lightweight neural network to perform local detection and tracking, offloading frames with low-confidence objects or high inter-frame deviation to the edge, and incorporated periodic feedback from the edge server to adjust thresholds dynamically. \cite{9488741} introduced an online learning algorithm to adaptively tune the motion deviation threshold for triggering detection. \cite{TMC_SongtaoGuo} further enhanced adaptability by jointly tuning a cumulative deviation threshold and frame resolution using contextual multi-armed bandit learning to optimize recognition accuracy and processing rate. Direct decision methods, on the other hand, determine the offloading strategy by solving an optimization problem over a batch of arriving frames, without explicitly managing frame queuing. For instance, \cite{ICASSP_MultiSever} addressed the offloading problem to multiple edge servers under minimum detection frequency constraints, aiming to maximize recognition accuracy while satisfying delay requirements. \cite{TMC_Min} proposed a link-adaptive scheme to jointly optimizes offloading decision and resolution selection.
% where only selected frames are detected, and intermediate frames are processed via local tracking.

Fixed thresholds simplify decision-making, but lack the adaptability required to handle dynamic or changing scenes. Adaptive thresholding provides greater flexibility and responsiveness but still struggles in scenarios involving high frame rates and computational load. Direct decision methods benefit from global optimization over a short-term observation window, yet they often overlook inter-window dependencies, which become critical under continuous and dynamic frame arrivals. In summary, existing approaches face significant limitations under heavy-load scenarios, due to the absence of frame queue management.

\subsection{Methods for Heavy-load Scenarios}
In heavy-load scenarios, frame queuing becomes a critical factor in decision-making for object tracking and detection \cite{QueueDropping,ChinaCom,Asy_MEC}. Similar to the light-load scenarios, the methods in the heavy-load scenarios can also be divided into threshold-based methods and direct decision methods. For example, a tile-level ``detect+track'' framework with adaptive tracker configuration was proposed in \cite{2021EdgeDuet}, where tracking priorities are dynamically updated to give precedence to high-speed objects.
% {\color{red}a fixed threshold was employed in \cite{2021EdgeDuet}, where object speed is estimated by continuously tracking motion across frames, and tracking priorities are dynamically updated to give precedence to high-speed objects using a single-object tracker.} 
For a more flexible tracking and detection switching, \cite{GLOBECOM_Xiaofeng} adopted the DRL to dynamically adjust both the deviation threshold and the set of frames to be processed within each decision epoch. 
% {\color{red}\cite{TWC_Offloading} formulated an optimization problem and made a direct decision on the frame resolution and server selection for object detection.}

As a specialized approach, parallel detection and tracking has been developed to handle heavy-load conditions and is categorized into intra-frame and inter-frame parallelism. Using inter-frame parallelism, \cite{9355581} developed a real-time system where detection and tracking operate concurrently across frames: while one frame undergoes detection, a lightweight tracker handles subsequent frames arriving during the detection interval. Intra-frame parallelism has been adopted by works such as \cite{10487981} and \cite{9796984}. In detail, \cite{10487981} proposed an on-device system for high-resolution videos that performs tracking across most regions within a single frame, invoking detection only in areas where tracking is unreliable (e.g., due to insufficient feature points). \cite{9796984} introduced a tracking-aware patching strategy that identifies regions likely to suffer from tracking failure and compacts them for detection. Naturally, intra-frame parallelism can be integrated into the inter-frame parallelism to enable faster video object recognition. 

A key limitation of these parallel strategies \cite{9355581, 10487981, 9796984} is that they require both a detector and a tracker to be deployed simultaneously on the device, which imposes significant hardware and energy demands, limiting their suitability for resource-constrained environments. Meanwhile, although the DRL-based adaptive threshold approach in \cite{GLOBECOM_Xiaofeng} demonstrates strong performance under high-load conditions, incorporating threshold selection into the decision space significantly expands the action set, resulting in high training complexity and learning overhead.

% Alternatively, parallel detection and tracking approaches have been developed to handle heavy-load conditions, categorized as intra-frame and inter-frame parallelism. In the inter-frame setting, \cite{9355581} proposed a real-time video processing system in which object detection and tracking run in parallel across frames: while the detector processes one frame, a lightweight tracker handles subsequent frames that arrive during the detection interval. In contrast, \cite{10487981} introduced an on-device system for high-resolution video that performs intra-frame parallelism by tracking most regions within each frame while applying object detection only to regions with tracking errors (e.g., insufficient feature points). Similarly, \cite{9796984} presented a tracking-aware patching strategy that enables intra-frame parallel execution by first identifying regions where tracking may fail and then packing those into a compact input for the detector.

% A common feature of \cite{9796984, 10487981, 9355581} is that all deploy both a detector and a tracker on the device simultaneously, which imposes high hardware demands and limits their applicability on resource-constrained devices. While the adaptive threshold-based decision-making approach in \cite{GLOBECOM_Xiaofeng} is efficient under heavy-load conditions, modeling the deviation threshold as part of the decision space results in a large action set, leading to high learning costs.

\subsection{Summary and This Work}
To enable efficient decision-making across both light- and heavy-load scenarios, we initiate our exploration in this paper. A preliminary version of this work has been accepted for presentation at IEEE/CIC ICCC 2025 \cite{ICCC_Yun}, where we focused on algorithm design in a single-device setting. In this extended version, we further explore a more realistic multi-device environment, present a generalized system model,  enhance the algorithm's generalization capability, and provide additional experimental results.

\section{System Model\label{sec: problem_description}}

\begin{figure}[t]
\centering
\includegraphics[width=0.48\textwidth]{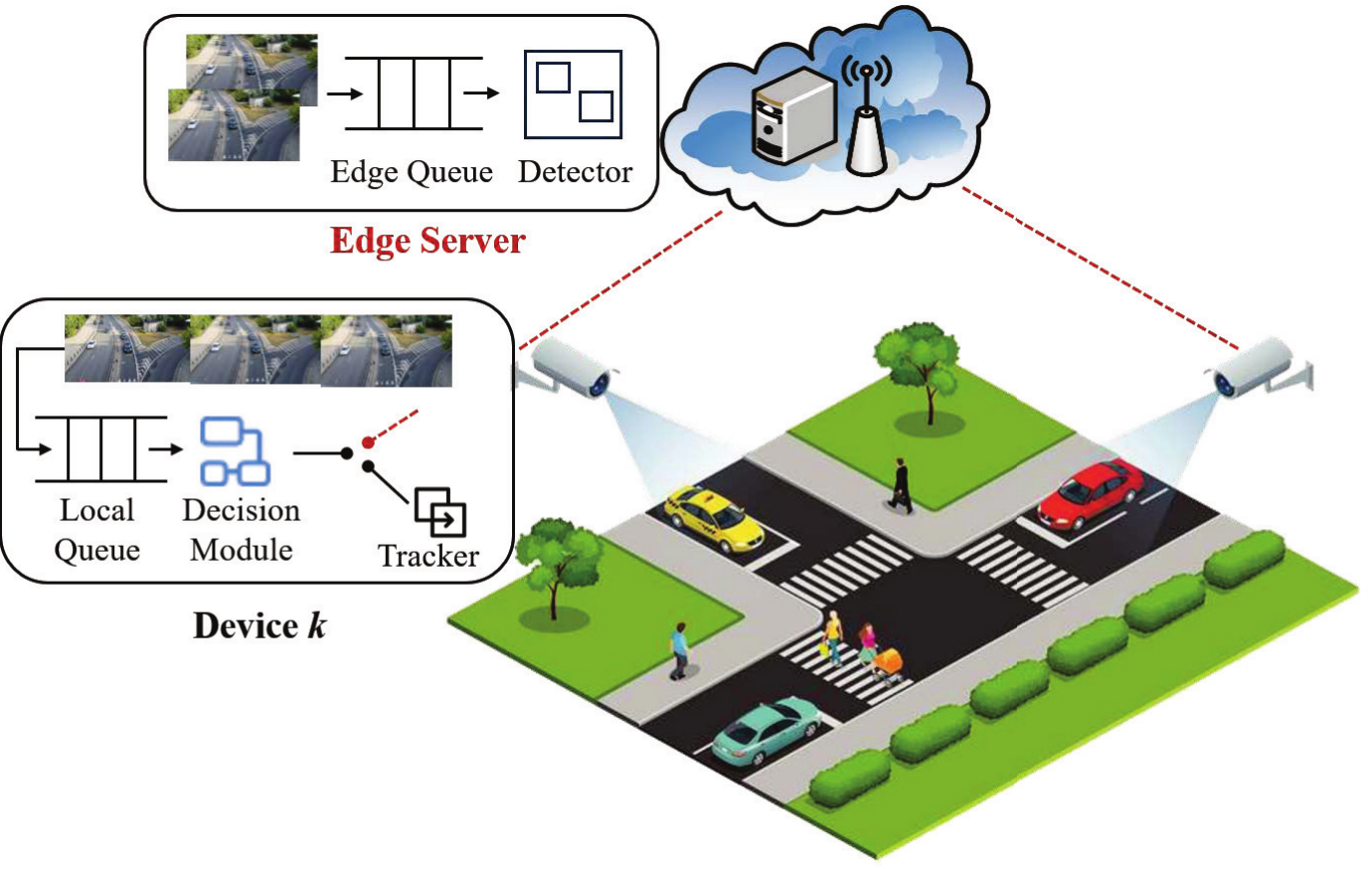}
\caption{An illustration of video object recognition in mobile edge networks.}
\label{fig:scenario}
\end{figure}

In this section, we elaborate on the arrival and object recognition of video frames, successively. Further, the related metrics are defined to evaluate the recognition performance.

\subsection{Frame Arrival}
As shown in Fig. \ref{fig:scenario}, we consider a mobile edge network, comprised of an access point connecting to an edge server and $K$ camera devices. Each device is able to capture and store the video frames. For high-efficiency video object recognition, we consider that the device with weak computing capability is responsible for the object tracking and the edge server with powerful computing capability is deployed with a large neural network model for object detection. Always, the device can flexibly configure frame rates to capture the video frame. The higher the frame rate is, the more smooth the video is but the higher the computational load is. Conversely, the lower the frame rate is, the less smooth the video is but the lower the computation load is. In this regard, the frame rate has a significant impact on the recognition performance.

Generally, we consider that a device $k\in\mathcal{K}\triangleq\{1,...,K\}$ captures a set of frames $ \mathcal{F}_k = \{1,...,F_k\}$ with frame rate $1/\Delta f_k$, which is in frames per second (FPS). Note that, $F_k$ and $\Delta f_k$ denote the number of frames and the capture interval between adjacent frames on device $k$. We assume that device $k$ and edge server are equipped with First-in-First-out (FIFO) queues to store the video frames \cite{C-RAN-JSAC,ChinaCom}, referred to as local queue $\mathcal{Q}_k$ and edge queue $\mathcal{Q}_0$, respectively. Specifically, $\mathcal{Q}_k$ on device $k$ is used to store its all frames waiting for object recognition and $\mathcal{Q}_0$ is used to store the frames from all devices for object detection. In this paper, we pay attention on the following two cases with the queue stability guaranteed:
\begin{itemize}
    \item Light-load case: In this case, $\Delta f_k$ is no smaller than the time for local tracking and that for edge detection in one frame. That is, there is no frames in queue $\mathcal{Q}_k$ waiting for the object recognition on device $k$. 
    \item Heavy-load case: In this case, $\Delta f_k$ is no smaller than the time for local tracking in one frame, but no larger than the time for edge detection in one frame. That is, queue $\mathcal{Q}_k$ becomes loaded when some frames are processed with edge detection, and remains empty when all frames (except for the initial one) are processed with local tracking.
\end{itemize}
 For a more smaller $\Delta f_k$, frames may be stacked in queue $Q_k$, thereby breaking the queue stability. To deal with this case, it is inevitable to consider the admission control scheme, which is out of our scope \cite{ChinaCom,Admission_Control}.

\begin{figure*}[tbh]
    \centering    \includegraphics[width=\textwidth]{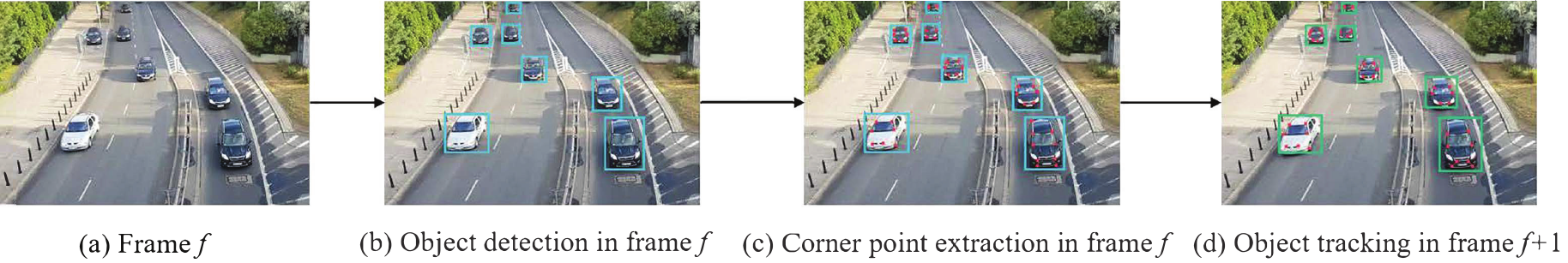}
    \caption{An illustration of object detection and tracking in frames.}
    \label{fig:detecting and tracking}
\end{figure*}

\subsection{Object Recognition}
It is observed from Fig. \ref{fig:scenario} that a general device $k$ is equipped with a decision module to determine where to process the head frame in its buffer queue. Particularly, we introduce $a_{k,f}\in\{0,1\}$ to indicate whether device $k$ tracks the objects in frame $f$ by itself (i.e., $a_{k,f}=1$) or detects the objects in frame $f$ with the assistance of edge server (i.e., $a_{k,f}=0$). An illustration of object detection and tracking process is presented in Fig. \ref{fig:detecting and tracking} and described in the following.

\subsubsection{Object Detection}
The device sends the frame to edge server, which is deployed with a pre-trained neural network model {\color{black}(e.g., Faster R-CNN \cite{ren2016fasterrcnnrealtimeobject} and YOLO \cite{CVPR_YOLO})} for object detection. After the edge server finishes object detection, the recognition results (e.g., object classes) and bounding boxes are obtained and returned back to the device. As shown in Fig. \ref{fig:detecting and tracking}(b), there may be multiple bounding boxes after object detection in frame $f$ of device $k$, whose set is expressed as $\mathcal{B}_{k,f}^\mathrm{D} = \{1,...,B_{k,f}^{\rm{D}}\}$ and corresponding ground truth is denoted by the set $\mathcal{B}_{k,f}^\mathrm{G} = \{1,...,B_{k,f}^{\rm{G}}\}$. For any bounding box $b\in\mathcal{B}_{k,f}^{\rm{D}}$, we can calculate the Intersection Over Union (IoU) between it and its ground truth $b^*\in\mathcal{B}_{k,f}^{\rm{G}}$ as follows:
\begin{equation}\label{eq-iou}
    {\rm{IoU}}_{b,b^*} = \frac{{\rm{Overlap}}(b,b^*)}{{\rm{Union}}(b,b^*)},
\end{equation}
where ${\rm{Overlap}}(b,b^*)$ and ${\rm{Union}}(b,b^*)$ represent the overlapped area and the union area between boxes $b$ and $b^*$. 
% These two areas can be calculated using vertex coordinates of boxes $b$ and $b^*$. 
Further, the mean IoU achieved by the object detection in frame $f$ of device $k$, is written as: 
\begin{equation}\label{eq: DmIoU}
    {\rm{mIoU}}_{k,f}^{\rm{D}} = \frac{\sum_{b\in\mathcal{B}_{k,f}^{\rm{D}}}{\rm{IoU}}_{b,b^*}}{B_{k,f}^{\rm{G}}}.
\end{equation}
Note that, we adopt the number of bounding boxes in ground truth set, i.e., $B_{k,f}^{\rm{G}}$, for the mean IoU, due to the fact that some objects maybe missed through the object detection.

\subsubsection{Object Tracking}
For convenience, we refer the detection frame to as the keyframe, which is used for object tracking in the subsequent frames. As shown in Fig. \ref{fig:detecting and tracking}(c) and (d), the device firstly finds out the corner points from the detected bounding boxes of the previous keyframe, and then tracks these corner points in the current frame, using some typical methods (e.g., optical flow \cite{Lucas1981AnII} and kernelized correlation filters \cite{KCF_Tracking}), to output the bounding boxes. Corner points are pixels with unique local structures in an image, usually locating at the intersections of image edges or at the locations with significant texture changes. These points exhibit good stability in image transformation and are therefore suitable for tracking\cite{harris1988combined}. Denote by $\mathcal{B}_{k,f}^{\rm{T}}=\{1,...,B_{k,f}^{\rm{T}}\}$ the set of bounding boxes from the object tracking in frame $f$ of device $k$. After the object tracking, the mean IoU for the corresponding frame is thus calculated as 
\begin{equation}
    {\rm{mIoU}}_{k,f}^{\rm{T}} = \frac{\sum_{b\in\mathcal{B}_{k,f}^{\rm{T}}}{\rm{IoU}}_{b,b^*}}{B_{k,f}^{\rm{G}}}.
    \label{eq: TmIoU}
\end{equation}

\subsection{Recognition Metrics}
To measure the performance of object recognition, we adopt the recognition accuracy, handling delay, and waiting delay as the metrics. The detailed definitions include several notations, which are summarized in Table \ref{tab: notations} for convenience, and are given in the following.
\begin{table}
\caption{The notation definitions for frame $f$ of device $k$.}
\label{tab: notations}
\centering
\renewcommand\arraystretch{1.2}
\begin{tabular}{|c|c|}
% \toprule
\hline
Notations & Definitions \\
\hline
% \midrule
$A_{k,f}$ & Recognition accuracy\\
$\mathrm{mIoU}_{k,f}^\mathrm{D}$ & Mean IoU for object detection \\
$\mathrm{mIoU}_{k,f}^\mathrm{T}$ & Mean IoU for object tracking \\
\hline
% \midrule
$H_{k,f}$ & Handling delay\\
$d_{k,f}^{\rm{UT}}$ & Uplink transmission delay\\
$d_{k,f}^{\rm{DT}}$ & Downlink transmission delay\\
$d_{k,f}^\mathrm{D}$ & Detection delay\\
$d_{k,f}^\mathrm{T}$ & Tracking delay\\
\hline
% \midrule
$W_{k,f}$ & Waiting delay\\
$w_{k,f}^\mathrm{0}$ & Waiting delay in queue $\mathcal{Q}_{0}$\\
$\tau_{k,f}^{0}$ & Arrival time in queue $\mathcal{Q}_{0}$\\
$t_{k,f}^{\rm{D}}$ & Detection completion time on edge server\\
$w_{k,f}$ & Waiting delay in queue $\mathcal{Q}_{k}$\\
$\tau_{k,f}$ & Arrival time in queue $\mathcal{Q}_{k}$\\ 
$t_{k,f}^{\rm{T}}$ & Tracking completion time on device $k$\\
$T_{k,f}$ & Completion time\\
\hline
% \bottomrule
\end{tabular}
\end{table}

\subsubsection{Recognition Accuracy}
The mean IoU in \eqref{eq: DmIoU} and \eqref{eq: TmIoU} are in the range of 0 to 1, and thus are used to measure the detection accuracy and tracking accuracy, respectively. With the consideration of decision action $a_{k,f}$, we then define the recognition accuracy for frame $f$ of device $k$, by 
\begin{equation}\label{dqn-1} 
  A_{k,f} = (1 - a_{k,f}){\rm{mIoU}}_{k,f}^\mathrm{D} + a_{k,f}{\rm{mIoU}}_{k,f}^\mathrm{T}.
\end{equation}

\subsubsection{Handling Delay}
We define handling delay as the time required to process a frame directly, without any waiting. For frame $f$ of device $k$, it is expressed as
\begin{equation}\label{dqn-2}
    H_{k,f} = (1 - a_{k,f})(d_{k,f}^\mathrm{UT} + d_{k,f}^\mathrm{D} + d_{k,f}^\mathrm{DT} ) + a_{k,f}{d}_{k,f}^\mathrm{T}.
\end{equation}
Specifically, if local tracking is performed for the frame, the handling delay equals the tracking delay $d_{k,f}^{\rm{T}}$. Otherwise, edge detection is executed and the handling delay equals the sum of the detection delay $d_{k,f}^{\rm{D}}$ and the communication delay, which includes the uplink transmission delay $d_{k,f}^{\rm{UT}}$ for frame uploading and the downlink transmission delay $d_{k,f}^{\rm{DT}}$ for the detection result downloading.

\subsubsection{Waiting Delay}
The frame is processed one by one and thus, the waiting delay for frame $f$ of device $k$ is given by 
\begin{equation}\label{eq: waiting_delay}
   W_{k,f} = (1 - a_{k,f})(w_{k,f} + w_{k,f}^0) + a_{k,f}w_{k,f}, 
\end{equation}
where $w_{k,f}$ and $w_{k,f}^0$ are the waiting delay in queue $\mathcal{Q}_k$ and $\mathcal{Q}_0$, respectively. That is, if the frame is processed with local tracking, it only experiences waiting in queue $\mathcal{Q}_k$ and its waiting delay equals to $w_{k,f}$; otherwise, the frame goes through both queue $\mathcal{Q}_k$ and $\mathcal{Q}_0$, and its waiting delay is a sum of $w_{k,f}$ and $w_{k,f}^0$.

% the waiting delay equals the waiting delay in queue $\mathcal{Q}_k$, denoted $w_{k,f}$, if it is processed with local tracking. If processed with edge detection, the waiting delay is the sum of waiting delays in both queue $\mathcal{Q}_k$ and $\mathcal{Q}_0$, denoted $w_{k,f}+w_{k,f}^0$. 
Picking the calculation of $w_{k,f}$ as an example, we show its dependencies on the arrival time of frame $f$ in queue $\mathcal{Q}_k$, i.e., $\tau_{k,f}$, and the completion time of its previous frame, i.e., $T_{k,f-1}$, as follows:
\begin{itemize}
    \item If frame $f$ is the first one in queue $\mathcal{Q}_k$, it can be directly processed without any waiting delay. 
    \item If frame $f$ arrives in queue $\mathcal{Q}_k$ before its previous frame is completed, its waiting delay is equal to $T_{k,f-1}-\tau_{k,f}$, otherwise, is zero.
\end{itemize}
On this basis, $w_{k,f}$ is expressed as: 
\begin{equation}\label{eq: waiting_delay_local}
w_{k,f} =  
\begin{cases}  
0, & f = 1, \\  
\max(0, T_{k,f-1} - \tau_{k,f}), & f = 2, \ldots, F.
\end{cases}  
\end{equation} 
% which means that once the previous frame is completely processed and returned to device $k$ with recognition results, the arriving frame incurs no waiting delay in queue $\mathcal{Q}_k$. 
In a similar way, for frame $f$ of device $k$ processed with edge detection, its waiting delay in queue $\mathcal{Q}_0$ is given by
\begin{equation}\label{eq: waiting_delay_edge}
w_{k,f}^0 =  
\begin{cases}  
0, & \text{if being the first frame}, \\  
\max(0, t_{k^{\prime},f^{\prime}}^{\rm{D}} - \tau_{k,f}^0), & \text{otherwise},
\end{cases}  
\end{equation}
% That is, there is no waiting delay for one frame when it arrives at queue $\mathcal{Q}_0$ after the previous frame is fully processed. 
where $\tau_{k,f}^0$ is its arrival time in queue $\mathcal{Q}_0$ and $t_{k^{\prime},f^{\prime}}^{\rm{D}}$ is the detection completion time of its previous frame, labeled as frame $f^\prime$ of device $k^{\prime}$.

In \eqref{eq: waiting_delay_local} and \eqref{eq: waiting_delay_edge}, $\tau_{k,f}$ is known and the other times, including $T_{k,f}$, $\tau_{k,f}^0$, and $\tau_{k,f}^{\rm{D}}$, depend on the detection and tracking decision. In detail, the completion time for frame $f$ of device $k$ is given by
\begin{equation}
\label{eq:completion_time} 
    T_{k,f} = (1-a_{k,f})(t_{k,f}^{\rm{D}}+d_{k,f}^{\rm{DT}}) + a_{k,f}t_{k,f}^{\rm{T}},
\end{equation}
where $t_{k,f}^{\rm{T}}$ and $t_{k,f}^{\rm{D}}$ are the tracking completion time on device $k$ and detection completion time on edge server, respectively. Consequently, \eqref{eq:completion_time} means that if the frame is processed with local tracking, its completion time equals to its tracking completion time, otherwise, equal to a sum of the detection completion time and the downlink transmission delay of detection results. Moreover, since tracking relies on detection results, the first frame on each device must be processed using edge detection, after which subsequent frames can be handled with local tracking. Hence, for frame $f$ of device $k$, once it is processed with local tracking, its $t_{k,f}^{\rm{T}}$ is written as:
\begin{equation}
     t_{k,f}^{\rm{T}} = \max(T_{k,f-1}, \tau_{k,f}) + d_{k,f}^{\rm{T}}, f=2,...,F,
\end{equation}
where the first term represents the tracking start time, defined as the later of the frame arrival time and the completion time of its previous frame, and $d_{k,f}^{\rm{T}}$ denotes the tracking delay. Once the frame is processed with edge detection, its $t_{k,f}^{\rm{D}}$ has the following recursive form:
\begin{equation}\label{eq:detection_completion_time}
    t_{k,f}^{\rm{D}} =  
    \begin{cases}  
    \tau_{k,f}^0+d_{k,f}^{\rm{D}}, & \text{if being the first frame}, \\  
    \max(\tau_{k,f}^0, t_{k^{\prime},f^{\prime}}^{\rm{D}})+d_{k,f}^{\rm{D}}, & \text{otherwise},
    \end{cases}  
\end{equation}
which means that in queue $\mathcal{Q}_0$, only the first frame is processed immediately with edge detection without any waiting. For each subsequent frame, the detection start time is the later of its arrival time and the detection completion time of the previous frame. In \eqref{eq:detection_completion_time}, $d_{k,f}^{\rm{D}}$ denotes the detection delay and the frame arrival time in queue $\mathcal{Q}_0$ is given by  
\begin{equation}\label{eq: arrival_time_edge}
\tau_{k,f}^0 =  
\begin{cases}  
d_{k,f}^{\rm{UT}}, & f = 1,\\  
\max(T_{k,f-1}, \tau_{k,f}) + d_{k,f}^{\rm{UT}}, & f = 2, \ldots, F,
\end{cases}  
\end{equation} 
where $d_{k,f}^{\rm{UT}}$ denotes the uplink transmission delay. For the first frame in queue $\mathcal{Q}_k$, its arrival time in queue $\mathcal{Q}_0$ is equal to its uplink transmission delay. For each subsequent frame, the uplink transmission starts at the later of its arrival time in queue $\mathcal{Q}_k$ and the completion time of the previous frame. Its arrival time in queue $\mathcal{Q}_0$ is then the sum of the uplink transmission start time and the transmission delay. 

\section{Problem Formulation and Algorithm Design in Single-Device Scenario\label{sec: single-device-alg}}
In this section, we focus on a single-device scenario to emphasize the design of an objective recognition algorithm that adapts to varying frame rates and performance requirements. To this end, we first formulate a long-term optimization problem and then leverage the DRL to make judicious decisions on edge detection and local tracking.

\subsection{Problem Formulation}
To achieve a balance between the recognition accuracy and delay, we define the achievable reward after processing frame $f$ of device $k$ as follows: 
\begin{equation}\label{dqn-7}
    R_{k,f} = A_{k,f} - \mathrm{\alpha}_k H_{k,f} - \mathrm{\beta}_k W_{k,f},
\end{equation}
where positive $\alpha_k$ and $\beta_k$ are regarded as importance factors of the handling delay and waiting delay, respectively. The larger $\alpha_k$ (or $\beta_k$) is, the smaller the handling delay (or the waiting delay) is. Note that, we assign different importance factors to devices to reflect their distinct performance requirements. For instance, some devices may prioritize high recognition accuracy even at the expense of increased delay (e.g. detailed object classification), while others may emphasize low recognition delay over high accuracy (e.g., real-time object avoidance). 

For a general device $k$, its achievable average reward can thus be maximized by optimizing the following problem:
\begin{align}
    {\text{(P0) }} \max_{a_{k,f}} \quad & \frac{1}{F}\sum_{f=1}^F R_{k,f} \nonumber\\ 
    \text{s.t. } \quad &\text{C1: } a_{k,f}\in\left\{0,1 \right\}, \forall f, \nonumber
\end{align}
where $R_{k,f}$ varies across frames, due to time-varying communication conditions and fluctuating computing capability in mobile edge networks, as well as dynamic changes in scene content. In addition, C1 means that frame $f$ is either processed with local tracking (i.e., $a_{k,f} =1$) or edge detection (i.e., $a_{k,f} =0$). Problem (P0) falls in the category of long-term optimization problem, which can be addressed effectively with the DRL. 

\subsection{DRL-based Algorithm Design}

We rewrite problem (P0) as a Markov decision process (MDP), comprised of state, action, and reward. Specifically, when processing frame $f$, device $k$ as the agent first captures a state $s_{k,f}$ from the time-varying environment, then outputs an action $a_{k,f}$ based on the state, and finally attain a reward $r_{k,f}$ to guide the subsequent action decisions. The detailed definitions of $s_{k,f}$, $a_{k,f}$, and $r_{k,f}$ are given below.  
\begin{itemize}
    \item \textbf{State:} For frame $f$ of device $k$, the state is defined as
    \begin{equation}\label{eq: state_single}
        s_{k,f} = \left\{o_{k,f},l_{k,f},v_{k,f}\right\}.
    \end{equation}
    Therein, $o_{k,f}$ represents the pixel deviation between frame $f$ and its previous keyframe, described as:
    \begin{equation}
        o_{k,f} = \left\{o_{k,f}^{\rm{x}},o_{k,f}^{\rm{y}}\right\},
    \end{equation}
    where $o_{k,f}^{\rm{x}}$ and $o_{k,f}^{\rm{y}}$ are the average absolute offsets along the $x$-axis and $y$-axis, respectively, between the corner points in the detected bounding boxes of the keyframe and their corresponding points in frame $f$ of device $k$. 
    Besides, $l_{k,f}$ also includes two elements:  
    \begin{equation}\label{eq: state_his}
        l_{k,f} = \left\{l_{k,f}^{\rm{Finv}},l_{k,f}^{\rm{Qlen}}\right\},
    \end{equation}
    which represent the interval between frame $f$ of device $k$ and its previous keyframe, as well as the current queue length in $\mathcal{Q}_k$ excluding frame $f$ of device $k$. Last but not least, $v_{k,f}$ is given by 
    \begin{equation}\label{eq: state_rate}
        v_{k,f} = \left\{v_{k,f}^{\rm{U}},v_{k,f}^{\rm{D}}\right\},
    \end{equation}
    which refer to the average transmission rates for frame uploading and detected result downloading, respectively.
    
    \item \textbf{Action:} The action taken by device $k$ is to decide whether to perform local tracking on frame $f$ or offload it for edge detection. In accordance with C1 in problem (P0), the action is defined as
    \begin{equation}\label{eq: action}
        a_{k,f} \in \left\{0,1\right\}.
    \end{equation}
    \item \textbf{Reward:} With aligning with the objective function of problem (P0), the reward of device $k$ is defined as
    \begin{equation}\label{eq: reward}
        r_{k,f} = R_{k,f}.
    \end{equation}
\end{itemize}
Then, we adopt the classical Deep Q-Network (DQN) for algorithm design \cite{DBLP:journals/corr/MnihKSGAWR13}. The proposed algorithm, LTED-Ada, short for Local Tracking and Edge Detection with Adaptation, adaptively selects between local tracking and edge detection, according to the frame rates and performance requirements. The complete LTED-Ada is summarized in Algorithm 1, which includes two phases: DQN training and inference.
% Our proposed algorithm can flexibly choose from the \underline{l}ocal \underline{t}racking and \underline{e}dge \underline{d}etection for video objective recognition, \underline{ada}pting to the frame rates, and thus is referred to as LTED-Ada. The LETD-Ada includes two phases: DQN training and inference. 

\begin{algorithm}[t]
\caption{LTED-Ada in single-device scenario \label{alg:dqn}}
\begin{algorithmic}[1]
\STATE \textbf{Training phase:}
\STATE For a general device $k$, initialize $\bm{\theta}_{k}$ and $\bm{\theta}^-_{k} = \bm{\theta}_{k}$;
\FOR{episode $= 1,..., E$}
    \STATE Observe state $s_{k,1}$ and execute action $a_{k,1} = 0$; 
    \FOR{$f =2,...,F$}
        \STATE Observe state $s_{k,f}$ and execute action $a_{k,f}$ as \eqref{eq: greedy_policy}; 
        \STATE Obtain reward $r_{k,f}$ and next state $s_{k,f+1}$; 
        \STATE Store transition $(s_{k,f}, a_{k,f}, r_{k,f}, s_{k,f+1})$ in replay memory $\mathcal{D}_k$;
        \STATE Every $\kappa_1$ frames, randomly sample a mini-batch from replay memory $\mathcal{D}_k$ to update $\bm{\theta}_{k}$ using the Adam optimizer\footnotemark;
         \STATE Every $\kappa_2$ frames, update $\bm{\theta}^-_{k} = \bm{\theta}_{k}$;
    \ENDFOR
\ENDFOR
\STATE Output optimal $\bm{\theta}_{k}^*$;
\STATE \textbf{Inference phase:}
\STATE Observe state $s_{k,f}$ for any frame $f$ and output $a_{k,f} = \arg\max_{a_{k,f}\in\{0,1\}} Q(s_{k,f}, a_{k,f}; \bm{\theta}_{k}^*)$.
\end{algorithmic}
\end{algorithm}
\footnotetext{Note that, this step is triggered when the number of samples in replay memory $\mathcal{D}_k$ exceeds the mini-batch size. Once the replay memory $\mathcal{D}_k$ is full, newly generated sample is stored by randomly discarding an older sample.}

\begin{figure}[t]
\centering
\includegraphics[width=0.48\textwidth]{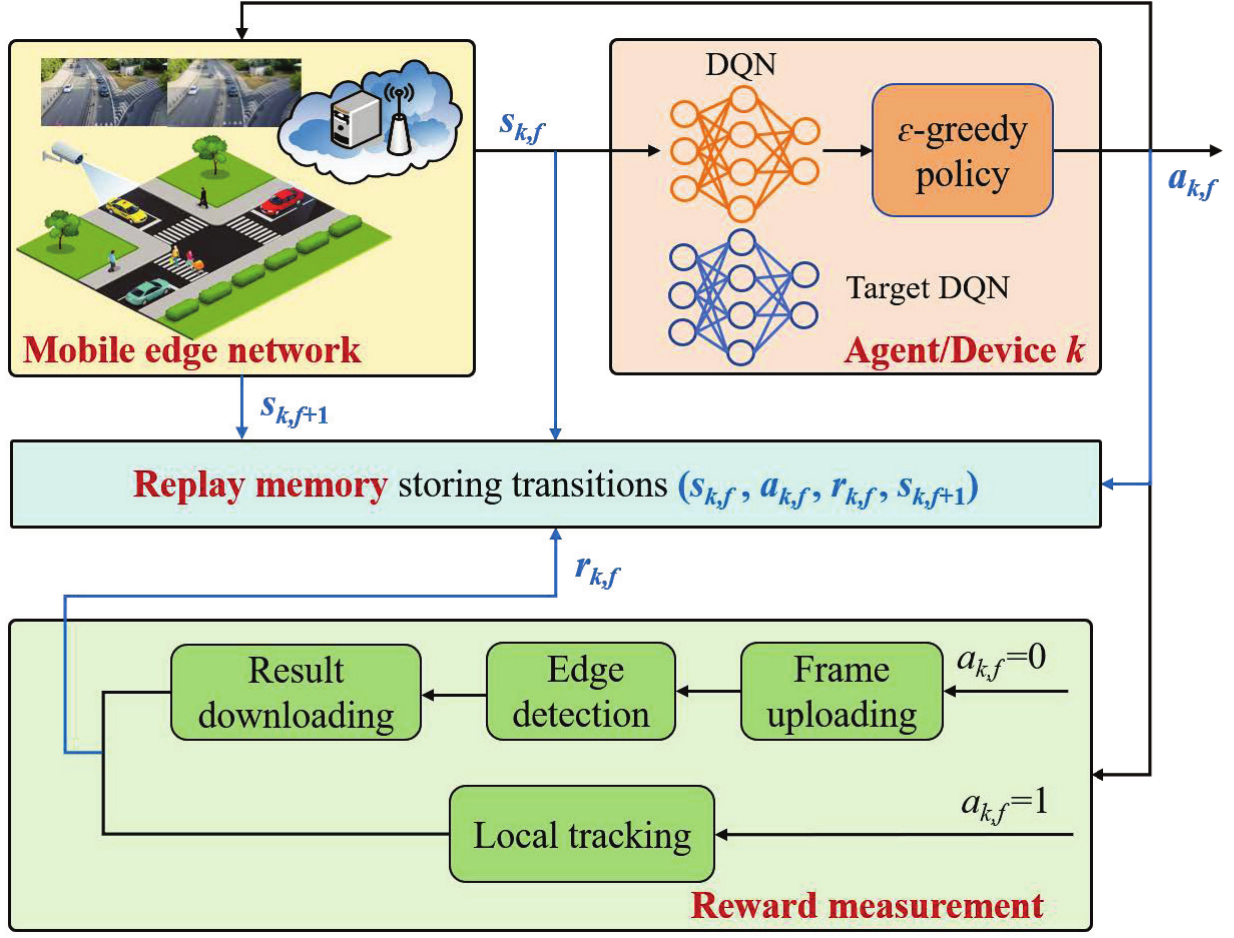}
\caption{Schematic diagram of DQN training in the proposed LTED-Ada.}
\label{fig:reinforcement}
\end{figure}

A sketch of the training phase is illustrated in Fig. \ref{fig:reinforcement}, in which a DQN with model parameter $\bm{\theta}_k$ is trained on device $k$ to approximate the state-value function $Q(s_{k,f},a_{k,f};\bm{\theta}_k)$. Further, device $k$ makes a decision for frame $f$ following the $\epsilon$-greedy policy:
\begin{equation}\label{eq: greedy_policy}
    a_{k,f}^* = \begin{cases}
        \text{Random selection between 0 and 1}, \text{with prob. } \epsilon\\
        \arg\max\limits_{a_{k,f}\in\{0,1\}} Q(s_{k,f},a_{k,f};\bm{\theta}_k), \text{ with prob. } 1-\epsilon.
    \end{cases}
\end{equation}
The other important point in the training phase is to update model parameter $\bm{\theta}_k$. To this end, a replay memory $\mathcal{D}_k$ is used to store samples from historical transitions. Specifically, a transition for frame $f$ of device $k$ is described as $(s_{k,f}, a_{k,f}, r_{k,f}, s_{k,f+1})$. Besides, a targe DQN with model parameter $\bm{\theta}_k^-$ is exploited to assist in $\bm{\theta}_k$ update. 

In detail, the loss function used for $\bm{\theta}_k$ update is defined as
\begin{equation}\label{eq: loss}
L_{k}(\bm{\theta}_k) = \mathbb{E}_{(s, a, r, s^{\prime})\in\mathcal{D}_k} \left[ \left( \hat{Q}(s,a) - Q(s, a; \bm{\theta}_k) \right)^2 \right],
\end{equation}
where $\hat{Q}(s,a)$ is given by
\begin{equation}\label{eq: Q_hat}
\hat{Q}(s,a) = 
\begin{cases}
r, & \text{if terminal} \\
r + \gamma \max\limits_{a^{\prime}\in\{0,1\}} Q(s^{\prime}, a^{\prime}; \bm{\theta}_k^-) , & \text{if not terminal}. 
\end{cases}
\end{equation}
For non-terminal states, $Q(s^{\prime}, a^{\prime}; \bm{\theta}_k^-)$ is the output of target DQN with state $s^{\prime}$ and action $a^{\prime}$, as well as, $\gamma\in[0,1)$ is a discount factor, which determines the significance of future reward relative to the immediate reward. If the current state is terminal (i.e., when $f=F$ in Algorithm 1), ${\hat{Q}(s,a)}$ is set to the immediate reward $r$ without considering the future reward.
The Adam optimizer is employed to update $\bm{\theta}_k$ with a learning rate of $\eta$.

In Algorithm 1, steps 2-13 outline the complete DQN training phase for the single-device scenario, which consists of $E$ episodes. In each episode, $F$ frames with labeled bounding boxes are used for the reward calculation and subsequently for the loss calculation in \eqref{eq: loss}. It is noted that, frame $1$, being the initial frame, must be processed with edge detection (as specified in step 4). Furthermore, the model parameter $\bm{\theta}_k$ is updated every $\kappa_1$ frames (as described in step 9) and synchronized to the target DQN model $\bm{\theta}_k^-$ every $\kappa_2$ frames (as described in step 10). Upon completion of the DQN training phase, the optimal model parameter $\bm{\theta}_k^*$ is achieved and used during the inference phase. As described in step 15, device $k$ first observes the state for the current frame, then leverages $\bm{\theta}_k^*$ to compute $Q$-values for all possible actions, and finally selects the action with the highest $Q$-value.

\section{Algorithm Extension to Multi-Device Scenario\label{sec: multi-device-alg}}
In the multi-device scenario, the LTED-Ada designed for the single device scenario (i.e., Algorithm 1) cannot be directly applied due to several limitations. First, the waiting delay in queue $\mathcal{Q}_0$ on the edge server becomes substantial when multiple devices simultaneously request edge detection. Second, as the device's requirements dynamically evolve to accommodate varying task demands, such as switching from real-time object avoidance to detailed object classification, the limited generalization capability of Algorithm 1 restricts its ability to make judicious decisions between local tracking and edge detection. 

To overcome these challenges, we tailor LTED-Ada for a multi-device setting with evolving frame rates and performance requirements, aiming to maximize the sum of average rewards across all devices, as follows:
\begin{align}
{\text{(P1) }} \max_{a_{k,f}} \quad & \frac{1}{F}\sum_{k=1}^K \sum_{f=1}^F R_{k,f} \nonumber \\
    \text{s.t. } \quad &  \text{C2: } a_{k,f}\in\left\{ 0,1 \right\},\forall k, \forall f. \nonumber
\end{align}  
Problem (P1) shares the same properties as problem (P0). Therefore, we reformulate it as an MDP and then apply the DRL to solve it as well.

In the multi-device scenario, the definitions of state, action, and reward for frame $f$ of device $k$ are given below:
\begin{itemize}
    \item \textbf{State:} The state is defined as
        \begin{equation}\label{eq: state_multi}
            s_{k,f} = \left\{o_{k,f},l^{\prime}_{k,f},v_{k,f}\right\},
        \end{equation}    
        with $l^{\prime}_{k,f}$ given by 
        \begin{equation}\label{eq: state_his_mul}
            l_{k,f}^{\prime} = \left\{l_{k,f}^{\rm{Finv}},l_{k,f}^{\rm{Qlen}}, l_{k,f}^{\rm{0,Qlen}}\right\},
        \end{equation}
        where $l_{k,f}^{\rm{Finv}}$ and $l_{k,f}^{\rm{Qlen}}$ have the same definitions as that in \eqref{eq: state_his}. Additionally, $l_{k,f}^{\rm{0,Qlen}}$ is introduced to indicate the length of queue $\mathcal{Q}_0$ on the edge server, when device $k$ starts to process frame $f$.
    \item \textbf{Action and reward:} The action and reward are the same as that defined in single-device scenario, following \eqref{eq: action} and \eqref{eq: reward}, respectively.
\end{itemize}
Then, we enhance LTED-Ada by incorporating a federated DQN training phase alongside a distributed inference phase. 

\begin{figure}[t]
\centering
\includegraphics[width=0.4\textwidth]{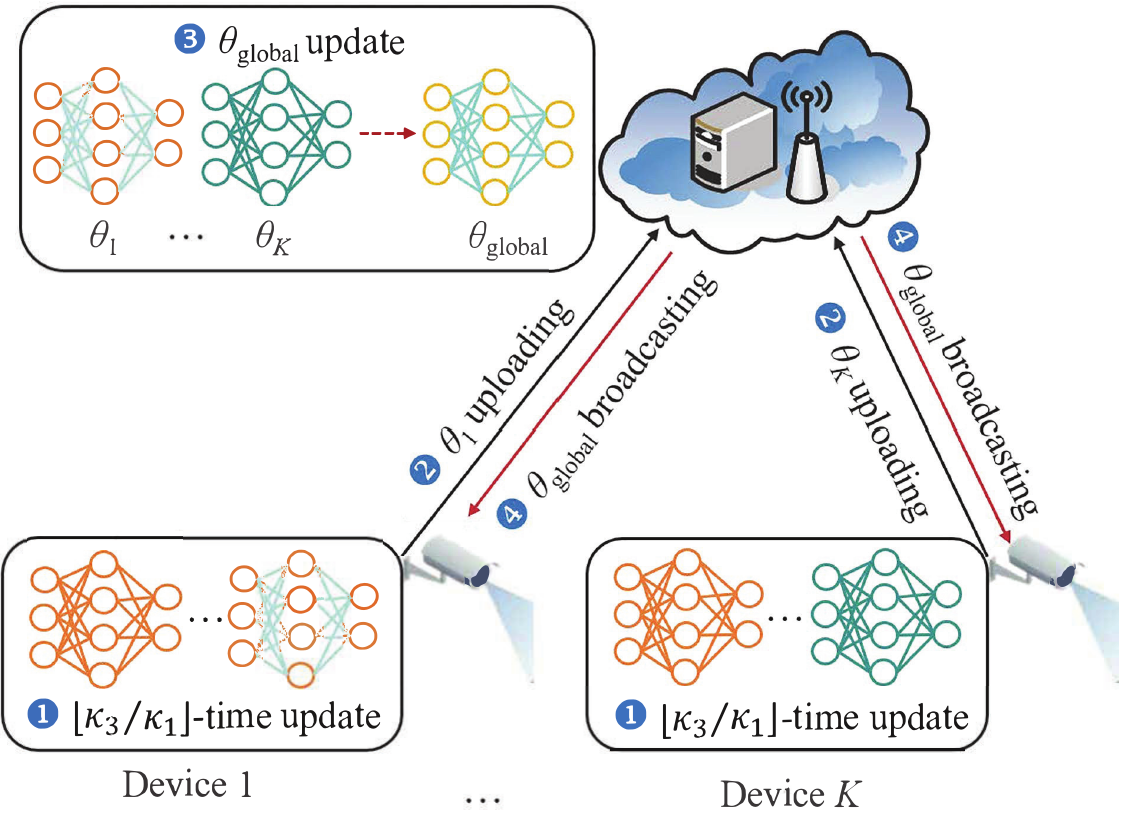}
\caption{{Federated training process in the proposed LETD-Ada.}}
\label{fig:fedDQN}
\end{figure}

The complete algorithm is summarized in Algorithm 2. During the training phase (steps 2-13), all devices train the same DQN model in parallel. For each device $k$, its DQN with model parameter $\bm{\theta}_k$ is trained locally, with the assistance of a target DQN with model parameter $\bm{\theta}_k^-$ in step 7. To enable $\bm{\theta}_k$ to adapt to varying frame rates and performance requirements, a key point lies in steps 8-9, where $\bm{\theta}_k$ is updated in a federated manner \cite{Kun_JSTSP}. Particularly, every $\kappa_3$ frames, device $k$ uploads $\bm{\theta}_k$ to the edge server for the global model update: 
\begin{equation}\label{eq: fl_model}
    \bm{\theta}_\mathrm{global} = \frac{1}{K}\sum_{k=1}^K \bm{\theta}_{k},
\end{equation}
which is then immediately distributed back to device $k$ for sequential $\bm{\theta}_k$ update. For clarity, the federated training process is illustrated in Fig. \ref{fig:fedDQN}. This approach allows the global model $\bm{\theta}_{\rm{global}}$ to capture the knowledge learned from multiple devices, enhancing the generalization capability of local model $\bm{\theta}_k$ at each device $k$. After the federated training phase terminates, the optimal global model  $\bm{\theta}_{\rm{global}}^*$ is attained and used for the distributed inference. As described in steps 15-17, each device $k$ updates its model parameter to $\bm{\theta}_k = \bm{\theta}_{\rm{global}}^*$ and selects the action corresponding to the maximum $Q$-value. 

\begin{algorithm}[t]
\caption{LTED-Ada in multi-device scenario}\label{alg:feddqn}
\begin{algorithmic}[1]
\STATE \textbf{Federated training phase:}\\
\STATE Initialize and broadcast $\bm{\theta}_{\rm{global}}$ to all devices; 
\FORALL{device $k$ in parallel}
    \FOR{episode $= 1,...,E$}
        \STATE Obtain state $s_{k,1}$ and execute action $a_{k,1} = 0$;
        \FOR{$f =2,...,F$}
            \STATE Invoke steps 6-10 in Algorithm 1; 
            \STATE Every $\kappa_3$ frames, device $k$ uploads $\bm{\theta}_k$ to the edge server to update $\bm{\theta}_{\rm{global}}$ as \eqref{eq: fl_model};
            \STATE The edge server immediately broadcasts the updated $\bm{\theta}_{\rm{global}}$, and device $k$ sets $\bm{\theta}_k=\bm{\theta}_{\rm{global}}$;
        \ENDFOR
    \ENDFOR
\ENDFOR
\STATE Output optimal $\bm{\theta}_{\rm{global}}^*$;
\STATE \textbf{Distributed inference phase:}
\FORALL{device $k$ in parallel}
    \STATE Obtain state $s_{k,f}$ for any frame $f$ and output $a_{k,f} = \arg\max_{a_{k,f}\in\{0,1\}} Q(s_{k,f}, a_{k,f}; \bm{\theta}_k^*)$ with $\bm{\theta}_k^* = \bm{\theta}_{\rm{global}}^*$.
\ENDFOR
\end{algorithmic}
\end{algorithm}

\section{Experimental Results\label{sec: results}}
In this section, we evaluate the performance of LTED-Ada in both single-device and multi-device scenarios through hardware-in-the-loop experiments. We first give the experimental setups, followed by extensive results and discussions.

\subsection{Experimental Setups}
\begin{figure}[tbp]
\centering
\includegraphics[width=0.45\textwidth]{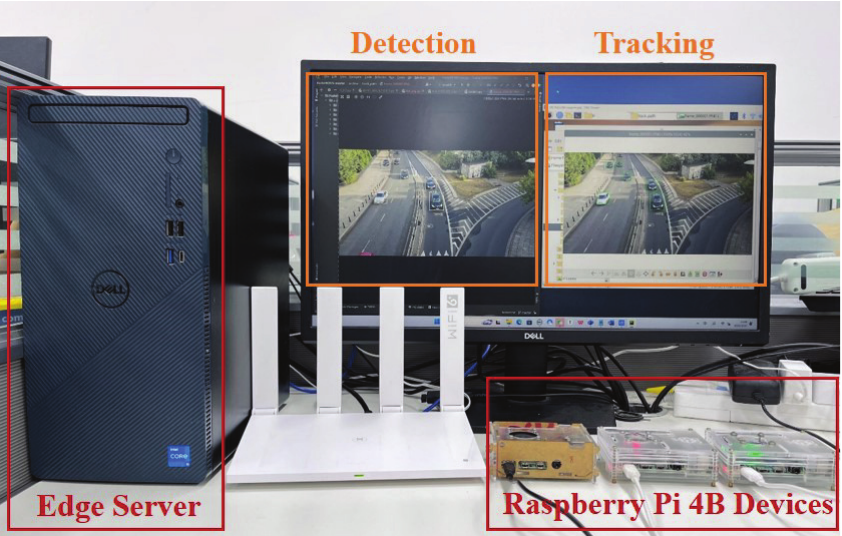}
\caption{\label{sce.png} Experimental setup with three Raspberry Pi 4B devices and a PC as the edge server.}
\end{figure}

To simulate a real mobile edge network, we use three Raspberry Pi 4B devices and a PC as the edge server, whose configurations are listed in Table \ref{tab:device_comparison}. All devices capture the same traffic video with 300 continuous frames \cite{10.1145/3387514.3405887} but with different capture intervals and diverse performance requirements on the recognition accuracy and delay. Each device is equipped with a decision modular to determine whether one frame is processed with edge detection or local tracking, using the proposed LTED-Ada. Moreover, the device employs the Lucas-Kanade optical flow algorithm for local tracking \cite{Lucas1981AnII}, with a per-frame tracking delay of approximately 0.47 seconds. The PC performs edge detection using a Faster R-CNN model with a ResNet-101 backbone \cite{ren2016fasterrcnnrealtimeobject}, incurring a per-frame detection delay of around 1.38 seconds. Wireless communication between the device and PC is established via 5 GHz Wi-Fi at a rate of 88.5 Mbps, resulting in an uplink transmission delay of 0.07 seconds and a negligible downlink delay. Fig. \ref{sce.png} shows our experimental setup, where a frame detected on the PC is displayed on the left side of the monitor and a frame tracked on the device is displayed on the right side through a remote desktop control tool, Virtual Network Computing (VNC) viewer.

\begin{table}[t]
\centering
\caption{\centering{Configurations of the PC and Raspberry Pi 4B.}}
\label{tab:device_comparison}

\begin{tabular}{ccc}
\toprule
Equipment & PC & Raspberry Pi 4B \\
\midrule
OS & 64-bit Windows 11  & Debian 10 (buster) \\
\addlinespace % Add some space between rows
CPU & Intel\textsuperscript{\textregistered} Core\textsuperscript{TM} i5-12400F & Quad-core Cortex-A72 \\
\addlinespace
GPU & GTX 1660 SUPER & VideoCore VI GPU \\
\addlinespace
Memory & 64 GB DDR4 & 4 GB LPDDR4 \\
\bottomrule
\end{tabular}
\end{table}

Unless otherwise specified, the default parameter settings for our proposed LTED-Ada are summarized in Table \ref{tab:example}. In detail, the trained DQN consists of three fully connected layers, with a hidden layer of 128 neurons. During training, an $\epsilon$-greedy policy is used for decision making. We initialize $\epsilon=\epsilon_0$ and update it each iteration according to $\epsilon=\epsilon*\epsilon_{\rm{decay}}$. Once $\epsilon$ reaches $\epsilon_{\min}$, it is set to $\epsilon = 0$. Additionally, the learning rate is initialized as $\eta = \eta_0$ and decayed every 10000 iterations by $\eta = \eta_0*\eta_{\rm{decay}}$. For comparisons, we adopt the following baselines: 
\begin{itemize}
    % \item Local tracking and detection (LT$\text{\&}$D): Both object tracking and detection are conducted at the local device without the assistance of edge server;
    \item Local tracking without detection (LTw/oD): Except for the first frame, all frames are processed at the device with local tracking;
    \item Edge detection without tracking (EDw/oT): All frames are offloaded to the edge server for object detection; 
    \item Local tracking and edge detection with fixed inter-frame interval (LTED-IntV): The frame is offloaded to the edge server for object detection with fixed inter-frame interval set to 15. During the interval, the arriving frames are processed at the device with local tracking;
    \item Local tracking and edge detection with fixed pixel deviation (LTED-DeV): Edge detection is performed for a frame when the pixel deviation between it and its previous keyframe is greater than a predefined threshold set to 10, otherwise local tracking is performed for this frame \cite{Glimpse}; 
    % {\color{purple}[The movement of pixels in the plane is represented by two dimensions, but the deviation represents the average value of the offset of pixels in the two-dimensional direction, which is a number.]}
    \item Random local tracking and edge detection (LTED-Rand): For a frame, local tracking or edge detection occurs, each with a probability of 0.5; 
    \item Parallel local tracking and edge detection (LTED-Paral): When a frame is processed with edge detection, the frames arriving during the detection of the last keyframe are processed concurrently with local tracking \cite{9355581};   
    \item Individual DQN based local tracking and edge detection (LTED-Indiv): Each device trains its own DQN to make object recognition decisions on its frames, without collaboration among devices.
\end{itemize}

\begin{table}[tbp]  % 尝试将表格放置在合适的位置
  \centering  % 使表格居中
  \caption{Algorithm parameter settings.}  % 先放置表格标题
  \label{tab:example}  % 表格标签，用于引用
  % 使用 minipage 控制表格宽度，例如宽度设为0.8\columnwidth
  \begin{minipage}{0.8\columnwidth}
    \centering
    \begin{tabular}{cc}  % 定义一个有三列的表格
      \toprule
      Parameter & Value \\  % 表头
      \midrule
      Hidden layers in DQN & 128 \\
      Discount factor $\gamma$ & 0.95\\
      Replay memory size & 10000\\
      Mini-Batch size & 64  \\ 
      $\kappa_1$, $\kappa_2$, $\kappa_3$  & 2, 100, 300  \\ 
      $\epsilon_{\rm{0}}$, $\epsilon_\mathrm{min}$, $\epsilon_\mathrm{decay}$ & 1, 0.001, 0.9999\\
      $\eta_{\rm{0}}$, $\eta_{\rm{decay}}$& 0.001, 0.1  \\  % 第三行数据 
      \bottomrule
    \end{tabular}
  \end{minipage}
\end{table}

We make comparisons with the consideration of different frame rates (determined by $\Delta f_k$) and performance requirements (characterized by $\alpha_k$ and $\beta_k$). By adjusting $(\Delta f_k,\alpha_k,\beta_k)$, we can get manifold combinations of load mode and performance requirement for device $k$. Note that, we use symbol '$/$' in the light-load mode, which can not only represent all possible $\Delta f_{k}$ making the waiting frames in queue $\mathcal{Q}_k$ empty, but also represent arbitrary $\beta_k$ due to the zero waiting delay. 
% Additionally, we evaluate performance based on the total recognition accuracy, total handling delay, total waiting delay, and total reward within one episode. 

\subsection{Performance of the Proposed Algorithm\label{sec: alg_performance}} 
In this subsection, we evaluate the convergence behavior and inference performance of the LTED-Ada in both single-device and multi-device scenarios.

\subsubsection{Single-Device Scenario}
\begin{figure}[t]
\centering
\includegraphics[width=0.5\textwidth]{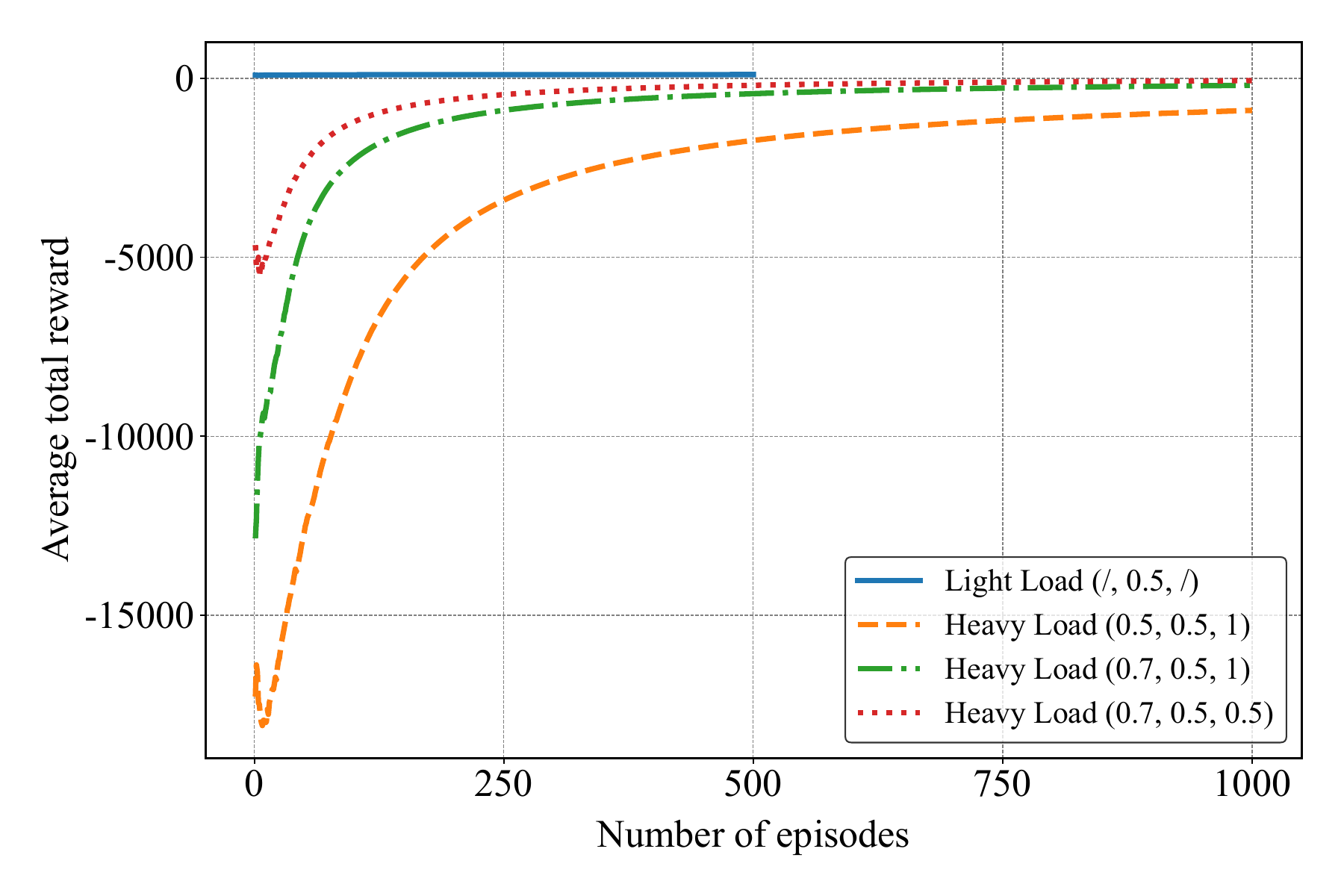}
\caption{\label{con_single}Convergence of the LTED-Ada in single-device scenario.} 
\end{figure}

Fig. \ref{con_single} shows the convergence of the proposed LTED-Ada, with $(\Delta{f}_k$, $\alpha_k$, $\beta_k)$ set as $(/, 0.5, /)$, $(0.5, 0.5, 1)$, $(0.7, 0.5 ,1)$, and $(0.7, 0.5, 0.5)$, respectively. We observe that the LTED-Ada faster converges to higher average total reward in light-load mode, due to the absence of complex queueing at the device. When the computational load is heavier (i.e., $\Delta f_{k}$ is smaller), the LTED-Ada converges more slowly to a lower average total reward. This is likely because more frames accumulate in the queue, leading to increased waiting delays, or more frames are processed using local tracking, which significantly reduces recognition accuracy. Additionally, as $\beta_k$ increases, the penalty for waiting delay becomes more severe, further hindering convergence and lowering the overall reward.

\begin{figure}[t]
\centering
\includegraphics[width=0.5\textwidth]{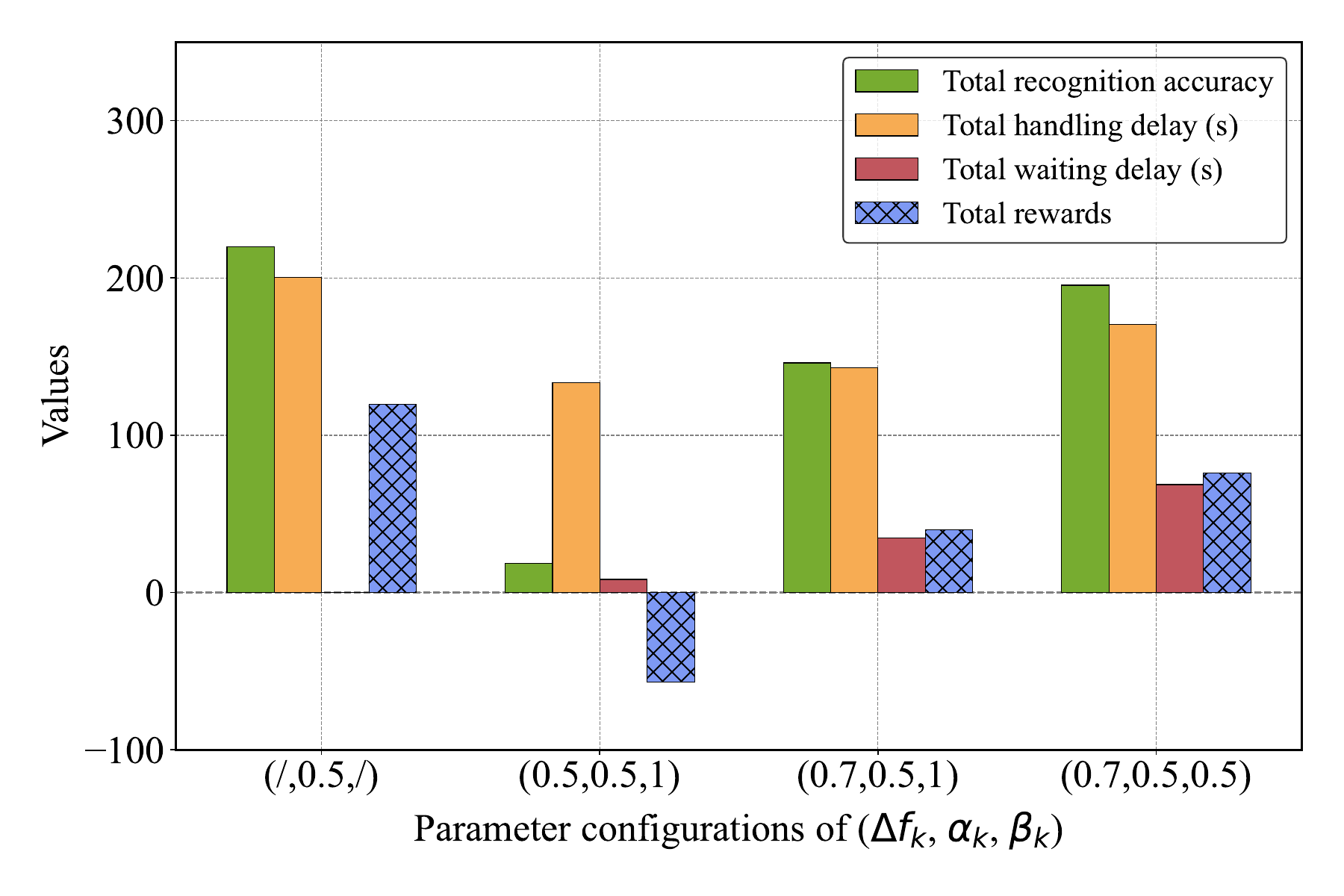}
\caption{\label{inf_sing}Performance comparisons of the LTED-Ada with different load modes and performance requirements, in single-device scenario.} 
\end{figure}

\begin{figure}[t]
\centering
\includegraphics[width=0.5\textwidth]{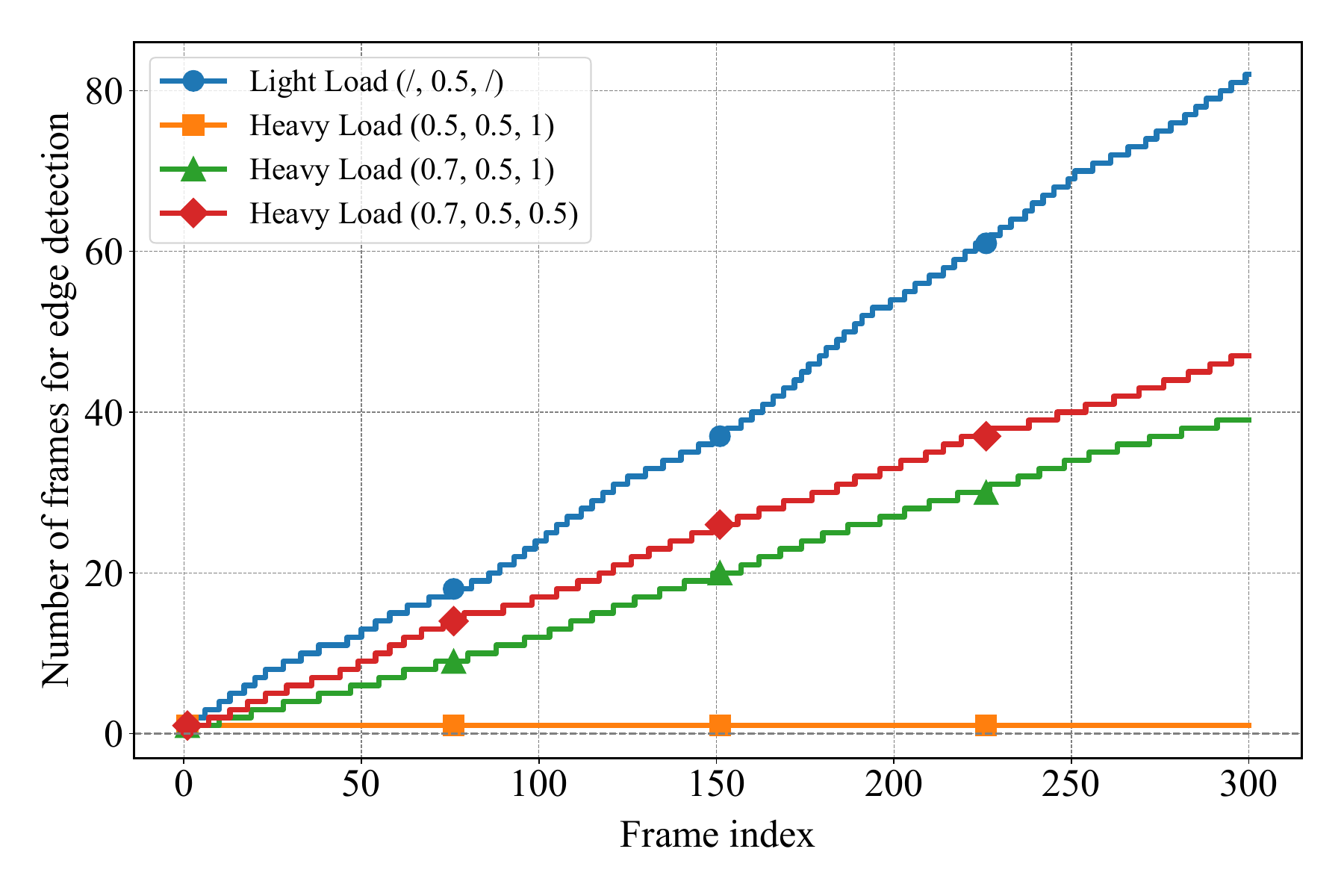}
\caption{\label{dec_sing}Decision results of the LTED-Ada with different load modes and performance requirements, in single-device scenario.} 
\end{figure}

Fig. \ref{inf_sing} shows the performance of the proposed LTED-Ada, under different $(\Delta{f}_k$, $\alpha_k$, $\beta_k)$ settings.
In the light-load mode with $(/, 0.5, /)$, the total recognition accuracy, handling delay, and reward are greater than that in the heavy-load mode with $(0.5, 0.5, 1)$, $(0.7, 0.5 ,1)$, and $(0.7, 0.5, 0.5)$. This is because, the LTED-Ada has zero waiting delay in the light-load mode, which gives a chance for more frames to be processed with edge detection. Through the comparisons in the heavy-load mode, it is confirmed that with smaller $\Delta f_{k}$, more frames are stacked on the device during the edge detection of one frame. Hence, to avoid exorbitant waiting delay, more frames are selected to be processed with local tracking, along with reduced recognition accuracy, handling delay, waiting delay, and reward. In addition, when $\beta_k$ becomes larger, the waiting delay will be reduced by tracking more frames. Even though, the impact of waiting delay on the reward is still significant, thereby resulting in worse reward. 

The underlying trends observed in Fig. \ref{inf_sing} can be further explained by Fig. \ref{dec_sing}, which presents the decision results of the LTED-Ada with different load modes and performance requirements. In the light load scenario with $(/, 0.5, /)$, the number of frames processed by edge detection is the highest, and the interval between two detection frames is the shortest. As $\Delta{f}_k$ decreases, the detection intervals become larger, and fewer frames are processed with edge detection. An extreme case occurs under heavy load with $(0.5, 0.5, 1)$, where all frames are processed using local tracking to reduce recognition delay at the expense of accuracy. 

\subsubsection{Multi-Device Scenario}
\begin{figure}[tbp]
\centering
\includegraphics[width=0.5\textwidth]{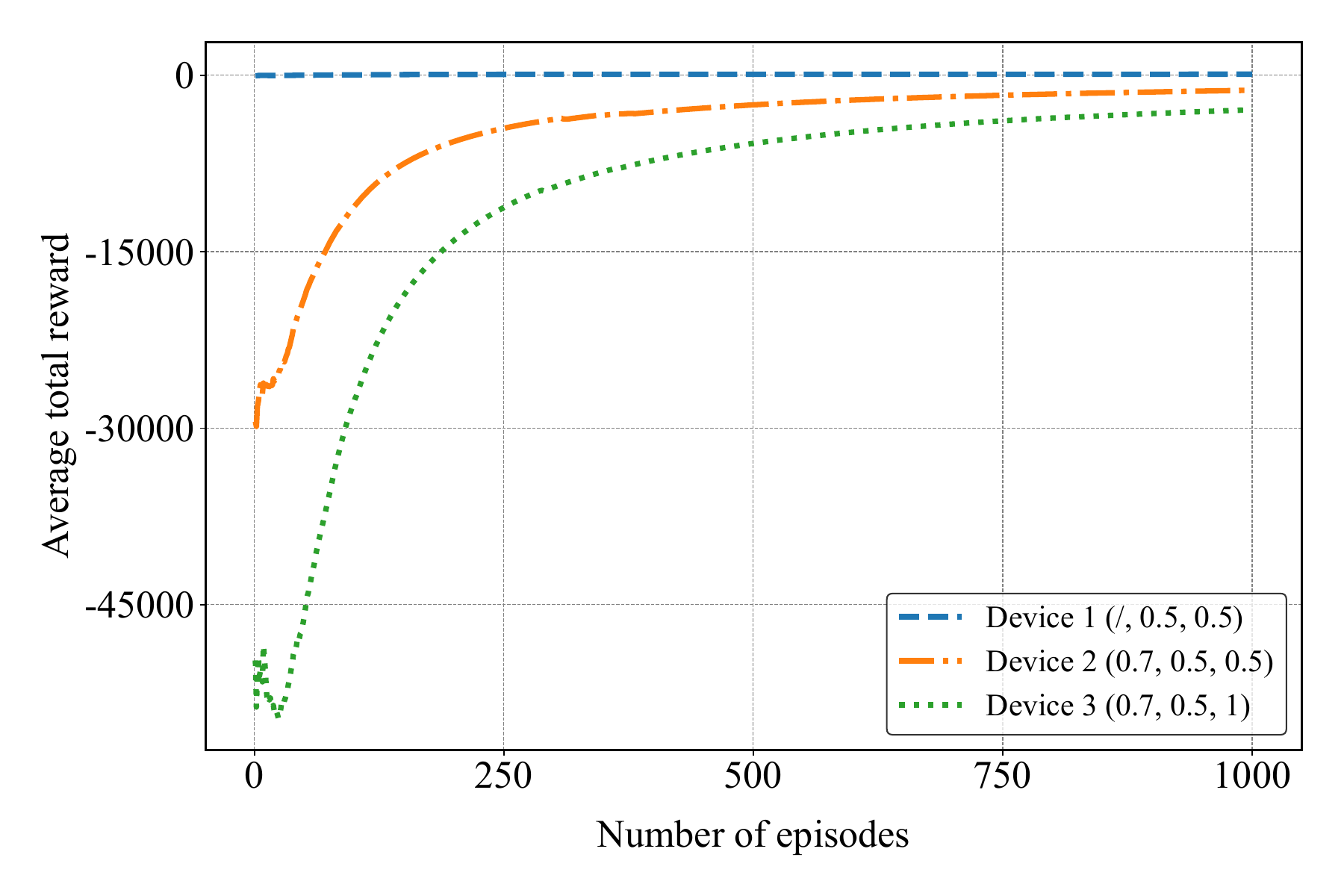}
\caption{\label{con_multi}Convergence of the LTED-Ada in multi-device scenario.} 
\end{figure}

In the multi-device scenario, we consider three devices with distinct load modes and performance requirements. As illustrated in Fig. \ref{con_multi}, device 1 is configured with $(/, 0.5, 0.5)$, device 2 with $(0.7, 0.5, 0.5)$, and device 3 with $(0.7, 0.5, 1)$ in the training phase. It can be seen that, the convergence performance of LTED-Ada in the multi-device setting closely aligned with that observed in the single-device case. Notably, device 1, operating under a light-load mode, quickly achieves the highest average total reward. In contrast, device 3, which imposes the strictest penalty on waiting delay, exhibits the lowest total reward. 

\begin{figure}[tbp]
\centering
\includegraphics[width=0.5\textwidth]{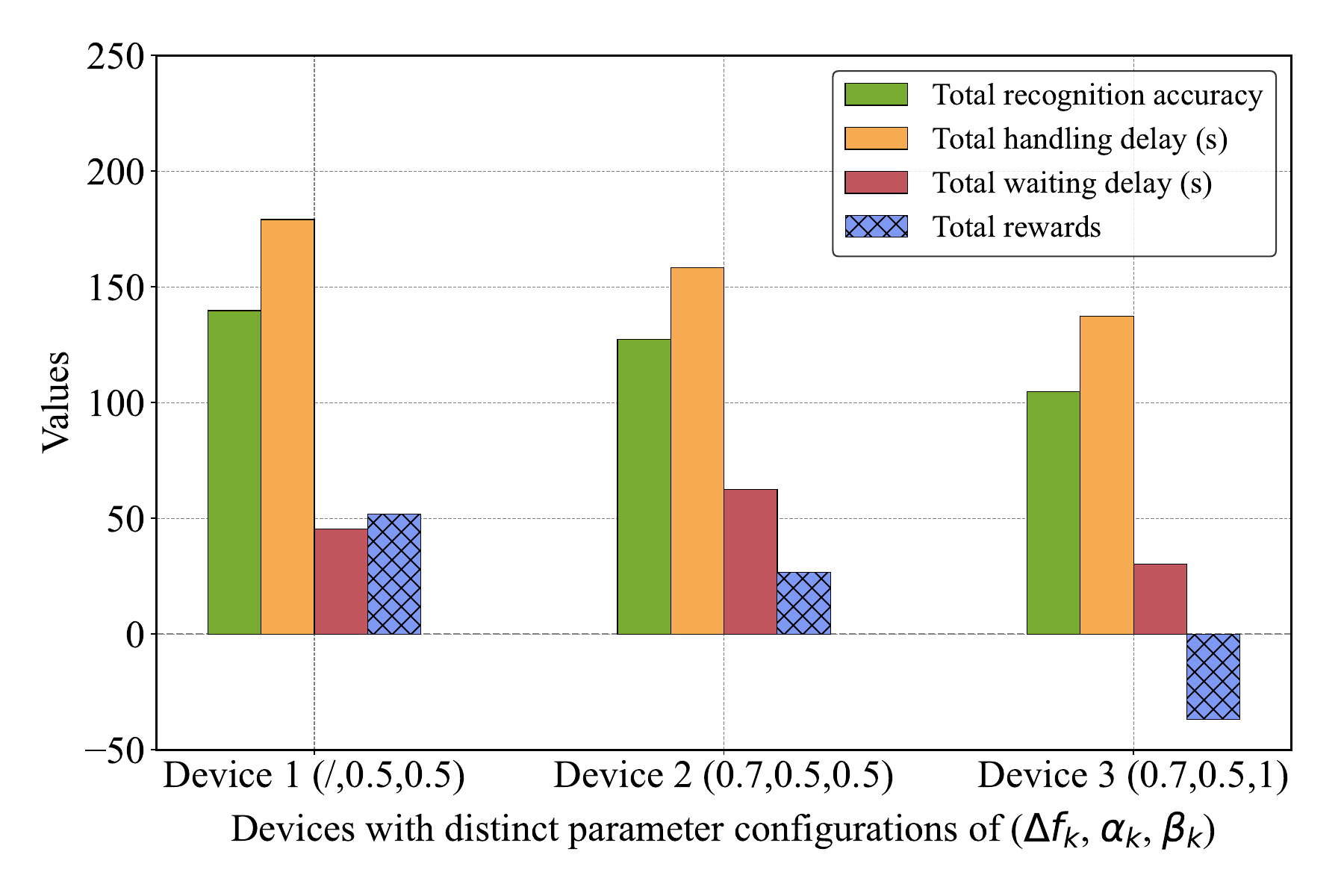}
\caption{\label{inf_multi}Performance comparisons among 3 devices using the LTED-Ada.} 
\end{figure}

% To validate the generalization capability of LTED-Ada with multi-device collaboration, 
During the inference phase, we consider 300 frames with varying frame rates and performance requirements. Specifically, the first 100 frames are set with $(/, 0.5, 0.5)$, the middle 100 frames with $(0.7, 0.5, 0.5)$, and the final 100 frames with $(0.7, 0.5, 1)$. In Fig. \ref{inf_multi}, we observe that device 1 achieves the highest total reward while device 3 achieves the lowest total reward. Notably, the total accuracy and total rewards are lower compared to the single-device results in Fig. \ref{inf_sing}, under the same parameter settings. This is because, the increase in the number of devices leads to a heavier load on the edge server. To reduce the waiting delay caused by the edge queuing, each device chooses to decrease the frequency of edge detection. 

\subsection{Performance Comparisons with Single Device}
In this subsection, we select appropriate baselines for comparison with our proposed LTED-Ada in single-device scenario under both light- and heavy-load modes. It is important to note that we specifically compare the baseline LTED-Paral with LTED-Ada in the heavy-load mode and exclude comparisons in the light-load mode, due to LTED-Paral's consideration of the waiting queue on the device.

\begin{figure}[tbp]
\centering
\includegraphics[width=0.5\textwidth]{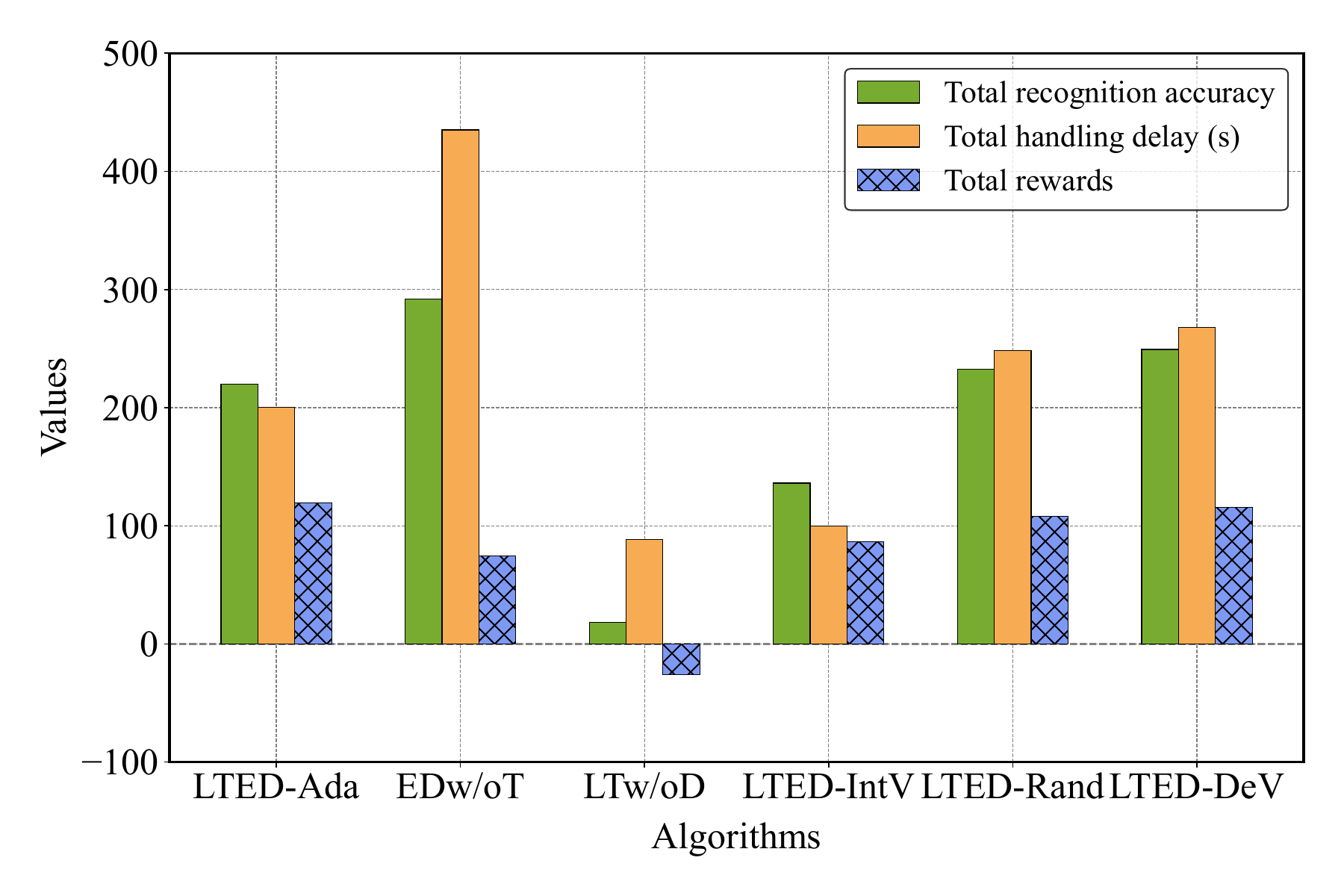}
\caption{\label{comp_light_sing}Performance comparisons in single-device and light-load scenario, with $(\Delta f_{k},\alpha_k,\beta_k) = (/, 0.5,/)$.} 
\end{figure}

Fig. \ref{comp_light_sing} shows the comparisons under light-load mode, with parameter $(\Delta f_{k},\alpha_k,\beta_k)$ set as $(/, 0.5,/)$. As expected, the LTED-Ada achieves the highest total reward. In comparison to EDw/oT and LTw/oD, LTED-Ada outperforms both due to its flexible selection between local tracking and edge detection. For the EDw/oT, it achieves the highest recognition accuracy, but results in a remarkable increase in handling delay. In contrast, LTw/oD exhibits the lowest handling delay, but significantly sacrifices recognition accuracy. Furthermore, the LTED-Ada can flexibly adjust the inter-frame interval for edge detection, giving it an advantage over the LTED-IntV and LTED-Dev. Finally, when compared with the LTED-Rand, the LTED-Ada performs better overall. The LTED-Rand, due to its suboptimal probability setting, processes more frames with edge detection, leading to improved recognition accuracy but deteriorated delay and even reward.

\begin{figure}[tbp]
\centering
\includegraphics[width=0.5\textwidth]{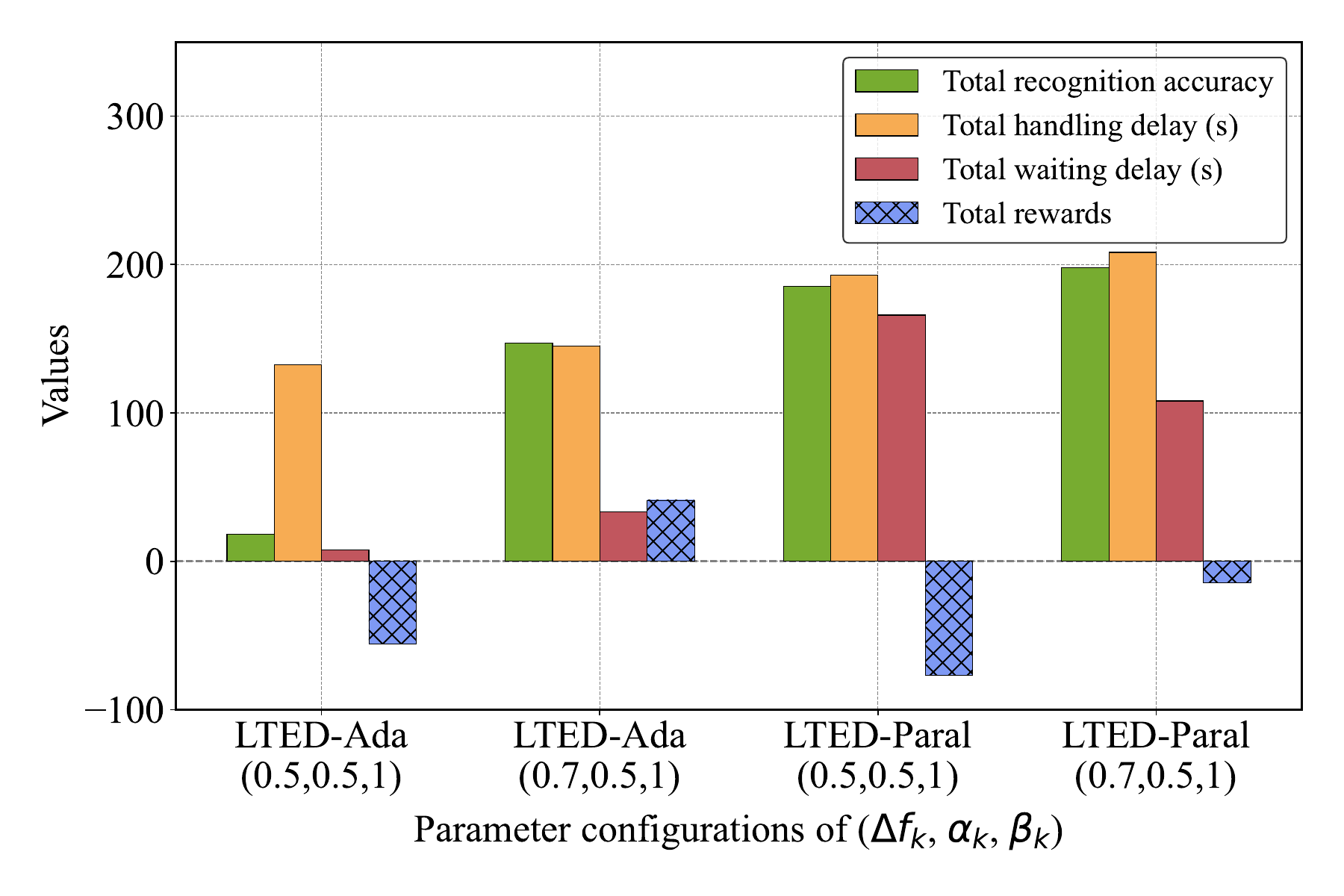}
\caption{\label{comp_heav_sing}{\color{black}Performance comparisons in single-device and heavy-load scenarios, with $(\Delta f_{k},\alpha_k,\beta_k)$ set as $(0.5, 0.5, 1)$ and $(0.7, 0.5, 1)$.}} 
\end{figure}

In Fig. \ref{comp_heav_sing}, we compare the LTED-Ada with LTED-Paral, where two heavy-load scenarios are considered with $(\Delta f_k, \alpha_k, \beta_k)$ set to $(0.5, 0.5, 1)$ and $(0.7, 0.5, 1)$, respectively.
In the LTED-Paral, the number of frames processed via local tracking is determined by $\Delta f_k$. As $\Delta f_k$ decreases (i.e., the load increases), more frames accumulate in the local queue and are processed locally, resulting in a slight improvement in total reward at the cost of reduced recognition accuracy.  This marginal gain highlights the limited adaptability of LTED-Paral to load variation.
In contrast, the proposed LTED-Ada responds to heavier loads by adaptively increasing the use of local tracking, significantly boosting total reward compared to the LTED-Paral.

\subsection{Performance Comparisons with Multiple Devices}

\begin{figure}[tbp]
\centering
\includegraphics[width=0.5\textwidth]{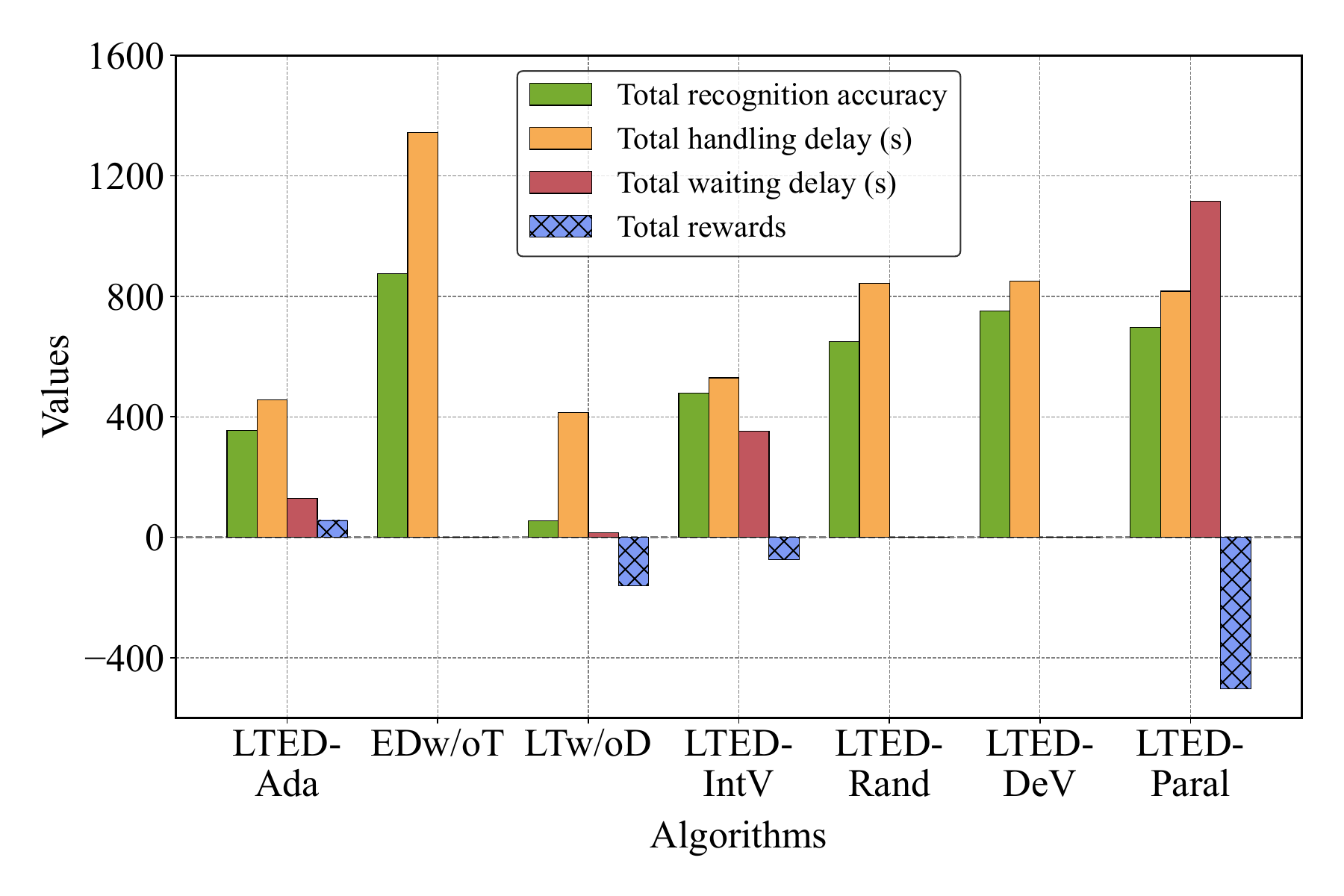}
\caption{\label{base_multi}{Performance comparisons of different algorithms with 3 devices, where $(\Delta f_{k}, \alpha_k, \beta_k)$ are set as $(/, 0.5, 0.5)$, $(0.7, 0.5, 0.5)$, and $(0.7, 0.5, 1)$, respectively, during the training phase.}}
\end{figure}

\begin{figure}[tbp]
\centering
\includegraphics[width=0.5\textwidth]{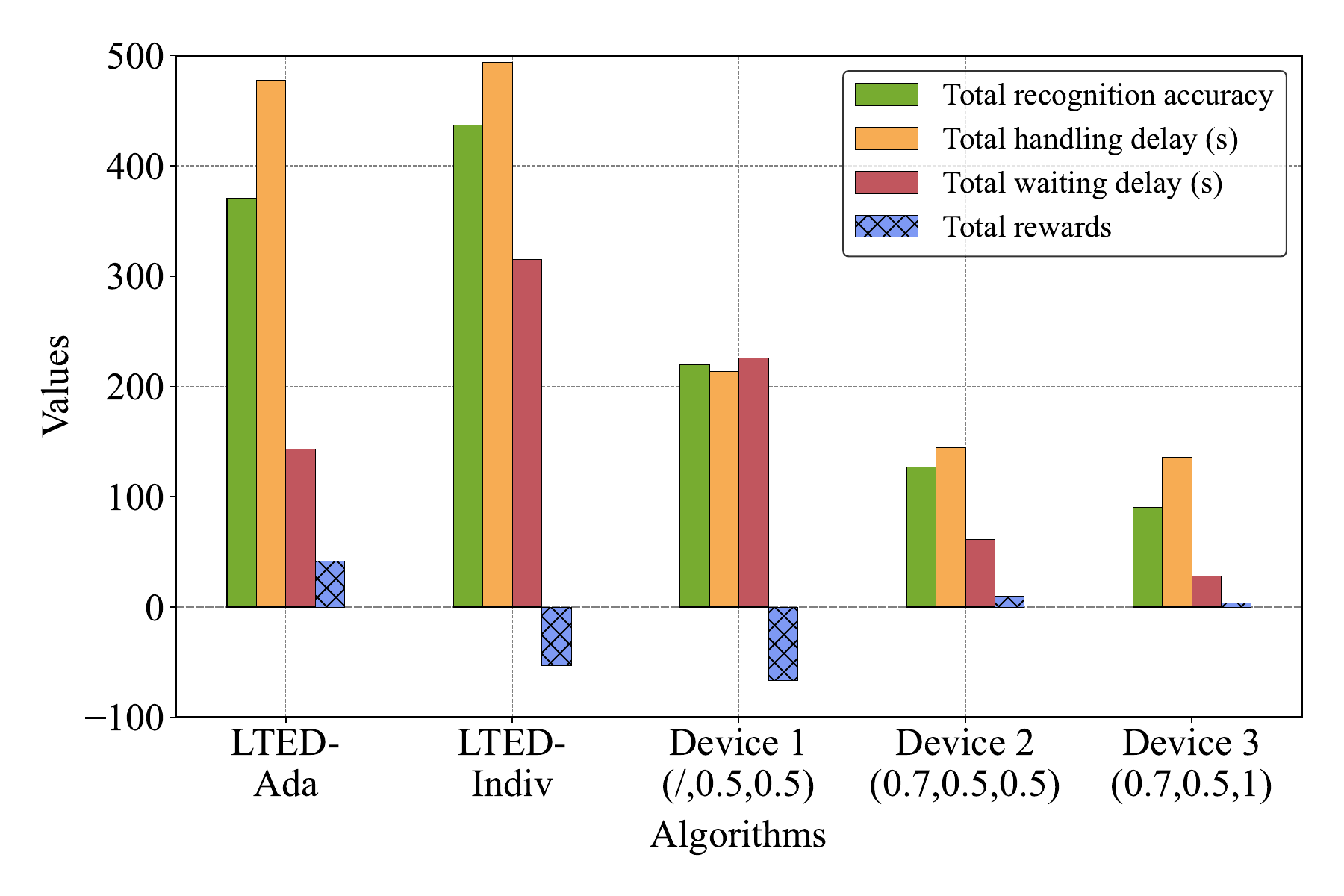}
\caption{\label{indiv_multi}{Performance comparisons between the LTED-Ada and LTED-Indiv with 3 devices, where $(\Delta f_{k}, \alpha_k, \beta_k)$ are set as $(/,0.5,0.5)$, $(0.7,0.5,0.5)$, and $(0.7,0.5,1)$, respectively, during the training phase.}} 

\end{figure}

Considering multiple devices, we compare the proposed LTED-Ada with all baseline algorithms previously used in single-device scenario. Additionally, to highlight the benefits of collaborative learning, we compare the LTED-Ada with LTED-Indiv, which trains DQN models independently for each device. During both the training and inference phases, all devices are configured using the same parameters as described in Section \ref{sec: alg_performance}.

As shown in Fig. \ref{base_multi}, the LTED-Ada achieves the highest reward by effectively balancing the recognition accuracy and delay. Note that, the EDw/oT, LTED-Rand, and LTED-DeV tend to process more frames using edge detection, leading to excessively high waiting delays and low rewards. To maintain clarity in the illustration, the results of these three schemes on the these two metrics are omitted. Compared to the LTw/oD, the LTED-Ada offers the flexibility to offload certain frames for edge detection, enhancing recognition accuracy and reward. Furthermore, the LTED-Ada processes more frames using local tracking than the LTED-IntV and LTED-Paral. Hence, it significantly reduces the recognition delay, particularly the waiting delay, while maintaining an acceptable level of recognition accuracy, ultimately leading to a higher overall reward.

We further compare the LTED-Ada with LTED-Indiv in Fig. \ref{indiv_multi}, along with the individual performance of the three devices under LTED-Indiv. For comparison, the individual performance of the three devices under LTED-Ada is shown in Fig. \ref{inf_multi}. By analyzing these two figures, we observe that the LTED-Ada achieves a higher total reward than LTED-Indiv overall. However, it does not consistently outperform the LTED-Indiv in terms of individual device rewards. For example, devices 1 and 2 achieve higher rewards with the LTED-Ada, while device 3 performs worse. This discrepancy is expected due to the averaging effect inherent in federated learning during the training phase of LTED-Ada.

\begin{figure}[tbp]
\centering
\includegraphics[width=0.5\textwidth]{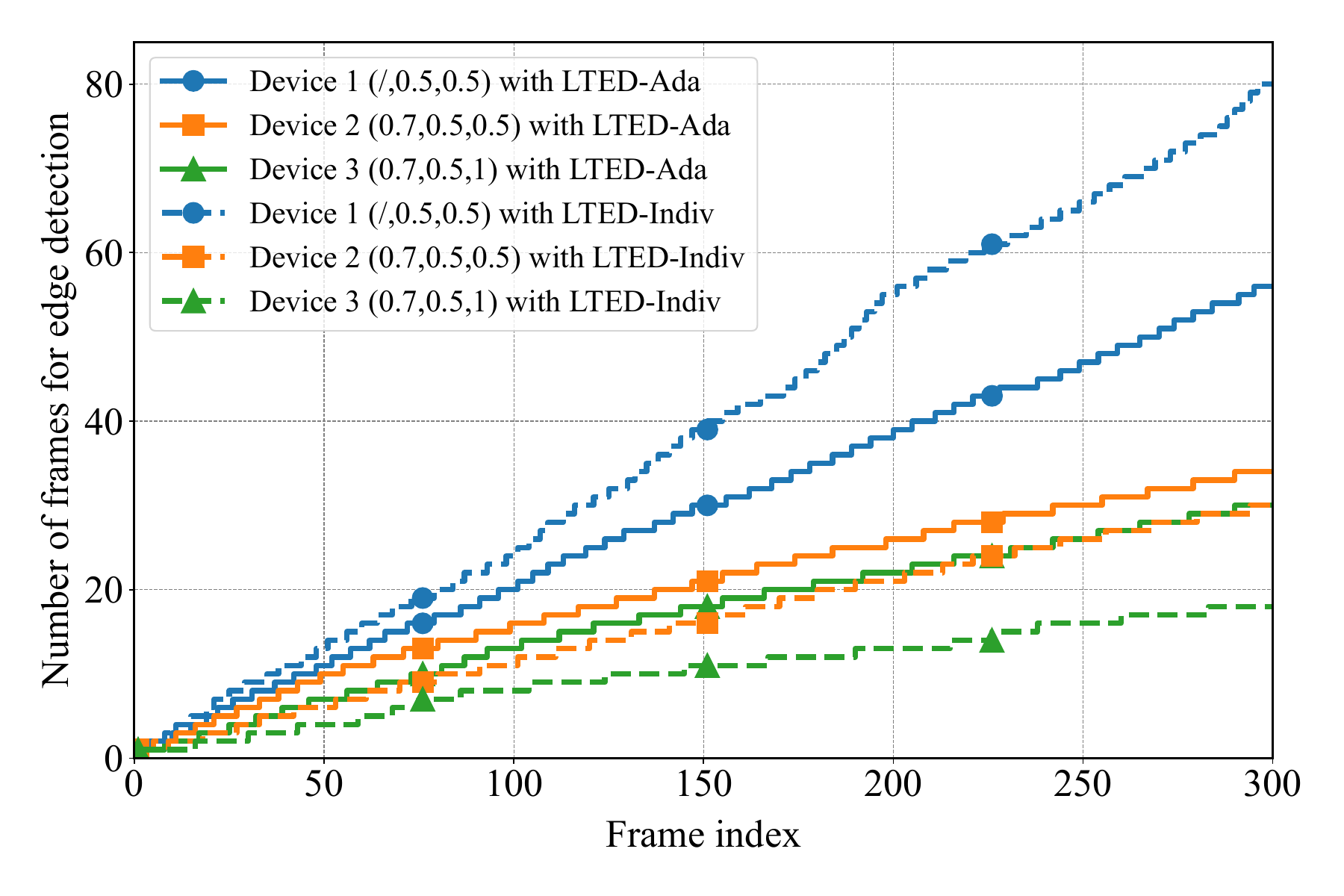}
\caption{\label{decison_multi}Decision results of the LTED-Ada and LTED-Indiv, with varying frame rates and performance requirements in multi-device scenario.} 
\end{figure}

To further illustrate the generalization capability of LTED-Ada, we present the decision results of both LTED-Ada and LTED-Indiv in Fig. \ref{decison_multi}. During the inference phase, each device processes 300 frames arriving with varying frame rates and performance requirements. Specifically, the first 100 frames are configured with $(/,0.5,0.5)$, the middle 100 with $(0.7,0.5,0.5)$, and the final 100 with $(0.7,0.5,1)$. Using LTED-Indiv as the baseline, we observe that device 1 significantly reduces the number of frames selected for edge detection in the last 200 frames under LTED-Ada, indicating an improved adaptability to heavier load due to knowledge shared from other devices. Device 2 increases the number of frames offloaded for edge detection in the first 100 frames and reduces it in the last 100 frames. Device 3 also selects more frames for edge detection in the first 200 frames than in the last 100. These results confirm that the LTED-Ada, through collaborative training among devices, better adapts to varying frame rates and performance requirements than LTED-Indiv.
 
\section{CONCLUSIONS\label{sec: conclusions}}

Considering the temporal correlation of consecutive frames and the dynamic conditions of mobile edge networks, we have proposed the LTED-Ada, a DRL-based video object recognition algorithm running on resource-constrained devices. The LTED-Ada intelligently selects between local tracking and edge detection, adapting to varying frame rates and performance requirements. Extensive hardware-in-the-loop experimental results have demonstrated that the proposed LTED-Ada outperforms multiple baselines in both single-device and multi-device scenarios, maximizing the total reward by effectively balancing recognition accuracy, handling delay, and waiting delay.

\bibliographystyle{IEEEtran}
\bibliography{video_recognition}
\end{document}